\documentclass{ecai}
\usepackage{times}
\usepackage{graphicx}
\usepackage{latexsym}

\usepackage{amsmath, amssymb, amsfonts, mathrsfs}
\usepackage{wrapfig, caption, multirow}
\usepackage{stmaryrd}
\usepackage[ruled,linesnumbered,noresetcount,vlined]{algorithm2e}
\usepackage{url}
\usepackage{picinpar}

\usepackage{color,soul} 

\SetAlgoSkip{}
\SetAlgoInsideSkip{}
\newboolean{includeMemo}
\newcommand{\memo}[1]{
  \ifthenelse {\boolean{includeMemo}}{\medskip\noindent\fbox{\begin{minipage}[b]{\dimexpr\linewidth-1em}#1\end{minipage}}\medskip\newline}}
\setboolean{includeMemo}{true} 
\newtheorem{lemma}{\bf Lemma}
\newtheorem{corollary}{\bf Corollary}

\newcommand{\bitemize}{\begin{list}{$\bullet$}{\topsep=1pt \parsep=0pt \itemsep=1pt \leftmargin=1em }} 
\newcommand{\eitemize}{\end{list}}

\DeclareMathOperator*{\bigtimes}{\vartimes}

\newcommand{\size}[1]{|#1|}

\def\is{\!=\!}
\def\st{\: | \:}

\newcommand{\argmax}{\operatornamewithlimits{argmax}}
\newcommand{\bemph}[1]{\textbf{\textit{#1}}}

\newcommand{\setf}[1]{{\bf{#1}}}
\newcommand{\varf}[1]{{\sl{{\small#1}}}}

\newcommand{\algf}[1]{{\sc{#1}}}
\newcommand{\procf}[1]{{\rm{#1}}}

\newcommand{\msgf}[1]{{\it{\sc{#1}}}}

\newcommand{\scope}[1]{\mathbf{x}^{#1}}

\newcommand{\oot}[1]{\multicolumn{#1}{c|}{--}}
\def\destroy{\circ}
\def\preserve{\star}
\def\ori{\diamond}
\def\LB{\textit{LB}}
\def\UB{\textit{UB}}
\def\minus{\text{-}}
\newenvironment{sproof}{\noindent {\underline{\it Proof (Sketch)}.}}{\hfill$\Box$}

\allowdisplaybreaks

\begin{document}

\title{Solving DCOPs with Distributed Large Neighborhood Search}

\author{Ferdinando Fioretto\institute{University of Michigan, Ann Arbor, US, email: fioretto@umich.edu} 
	   \and Agostino Dovier \institute{University of Udine, Udine, IT, email: agostino.dovier@uniud.it}
	   \and Enrico Pontelli\institute{New Mexico State University, Las Cruces, US, email: epontell@cs.nmsu.edu} 
	   \and William Yeoh\institute{New Mexico State University, Las Cruces, US, email: wyeoh@cs.nmsu.edu} 
	   \and Roie Zivan\institute{Ben Gurion University, Beersheba, IL, email: zivanr@cs.bgu.ac.il}}

\maketitle
\bibliographystyle{ecai}

\begin{abstract}
The field of Distributed Constraint Optimization has gained momentum in recent years, thanks to its ability to address various applications related to multi-agent cooperation. Nevertheless, solving Distributed Constraint Optimization Problems (DCOPs) optimally is NP-hard. Therefore, in large-scale, complex applications, incomplete DCOP algorithms are necessary. Current incomplete DCOP algorithms suffer of one or more of the following limitations: they 
{\bf (a)}~find local minima without providing quality guarantees; 
{\bf (b)}~provide loose quality assessment; or 
{\bf (c)}~are unable to benefit from the structure of the problem, such as domain-dependent knowledge and hard constraints. 
Therefore, capitalizing on strategies from the centralized constraint
solving  community, we propose a \emph{Distributed Large Neighborhood Search (D-LNS)} framework to solve DCOPs.\footnote{
An extended abstract of this work appeared in \cite{fioretto:15b}.
}
The proposed framework (with its novel repair phase) provides guarantees on solution quality, refining upper and lower bounds during the iterative process, and can exploit domain-dependent structures. Our experimental results show that D-LNS outperforms other incomplete DCOP algorithms on both structured and unstructured problem instances.
\end{abstract}

\section{Introduction}
In a \emph{Distributed Constraint Optimization Problem (DCOP)}, multiple
 agents coordinate their value assignments to maximize the sum of resulting constraint utilities~\cite{modi:05,yeoh:12}. DCOPs represent a powerful approach to the description and solution of many practical problems in 
 a variety of application domains, such as distributed scheduling, coordination of unmanned air vehicles, smart grid electrical networks, and sensor networks~\cite{ramchurn:11,zivan:14,kumar:09,stranders:09b}. 

In many cases, the coordination protocols required for the complete resolution of DCOPs demand a vast amount of resources and/or communication,  making them infeasible to solve real-world complex problems. In particular complete DCOP algorithms find  optimal  solutions at the cost of a large runtime or network load, while 
incomplete approaches trade optimality for lower usage of resources. Since finding optimal DCOP solutions is NP-hard, incomplete algorithms are often necessary to solve large interesting problems. Unfortunately, several local search algorithms (e.g.,~DSA~\cite{zhang:05}, MGM~\cite{maheswaran:04b}) and local inference algorithms (e.g.,~Max-Sum~\cite{farinelli:08}) do not provide  guarantees on the quality of the solutions found. More recent developments, such as region-optimal algorithms~\cite{pearce:07,vinyals:11}, Bounded Max-Sum~\cite{rogers:11}, and DaC algorithms~\cite{vinyals:10b,hatano:13} alleviate this limitation. 
Region-optimal algorithms allow us to specify regions with a maximum size of $k$ agents or $t$ hops from each agent, and they optimally solve the subproblem within each region. Solution quality bounds are provided as a function of $k$ and/or $t$. Bounded Max-Sum is an extension of Max-Sum, which solves optimally an acyclic version of the DCOP graph, bounding its solution quality as a function of the edges removed from the cyclic graph. DaC-based algorithms use Lagrangian decomposition techniques to solve agent subproblems sub-optimally. 
Good quality assessments are essential for sub-optimal  solutions. However, many  incomplete DCOP approaches can provide arbitrarily poor quality assessments (as confirmed in our experimental results). In addition, they are unable to exploit domain-dependent knowledge or the hard constraints present in problems. 

In this paper, we address these limitations by introducing the \emph{Distributed Large Neighborhood Search (D-LNS)} framework. D-LNS solves DCOPs by building on the strengths of \emph{centralized} LNS~\cite{shaw:98}, a \emph{centralized} meta-heuristic that iteratively explores complex neighborhoods of the search space to find better candidate solutions. LNS has been shown to be very effective in solving a number of optimization problems~\cite{godard:05,ropke:06}.
While typical LNS approaches focus on iteratively refining lower bounds of a solution, we propose a method that can iteratively refine both lower and upper bounds of a solution, imposing no restrictions (i.e., linearity or convexity) on the objective function and constraints.

This work advances the state of the art in DCOP resolution:
	{\bf (1)} We provide a novel distributed local search framework for DCOPs, which provides quality guarantees by refining both upper and lower bounds of the solution found during the iterative process;
	{\bf (2)} We introduce two novel distributed search algorithms, DPOP-DBR and T-DBR,  built within the D-LNS framework, and characterized by the ability to exploit problem structure and offer low network usage---T-DBR provides also a low computational complexity per agent; and
	{\bf (3)} Our evaluation against  representatives of search-based, inference-based, and region-optimal-based 
	incomplete DCOP algorithms shows that 
	T-DBR converges faster to better solutions, provides tighter solution quality bounds, and is more scalable.

The rest of the paper is organized as follows. In the next section, we introduce DCOPs and review centralized LNS. Section \ref{sec:dlns} presents our novel D-LNS schema. 
Section \ref{sec:DBR} presents a general algorithm framework, based on D-LNS, that iteratively refines lower and upper bounds of the DCOP solutions. We further describe two implementations of such framework offering different tradeoffs of agent complexity and solution quality. 
Prior concluding the Section, we report an example trace of the proposed repair algorithm, aimed at elucidate its behavior within the D-LNS framework.
Section \ref{sec:theo} discusses the theoretical properties of the algorithms presented, with particular emphasis on the correctness for the solution bounds returned during the iterative process.
We present the related works in Section \ref{sec:related}, and summarize our evaluation of the proposed framework against search-based, inference-based, and region-optimal-based DCOP incomplete algorithms, in Section \ref{sec:results}. Finally, Section \ref{sec:conclusions} concludes the paper.


\section{Background}\label{sec:DCOP}

\begin{figure}[!t]
	  \centering{\includegraphics[width=0.45\textwidth]{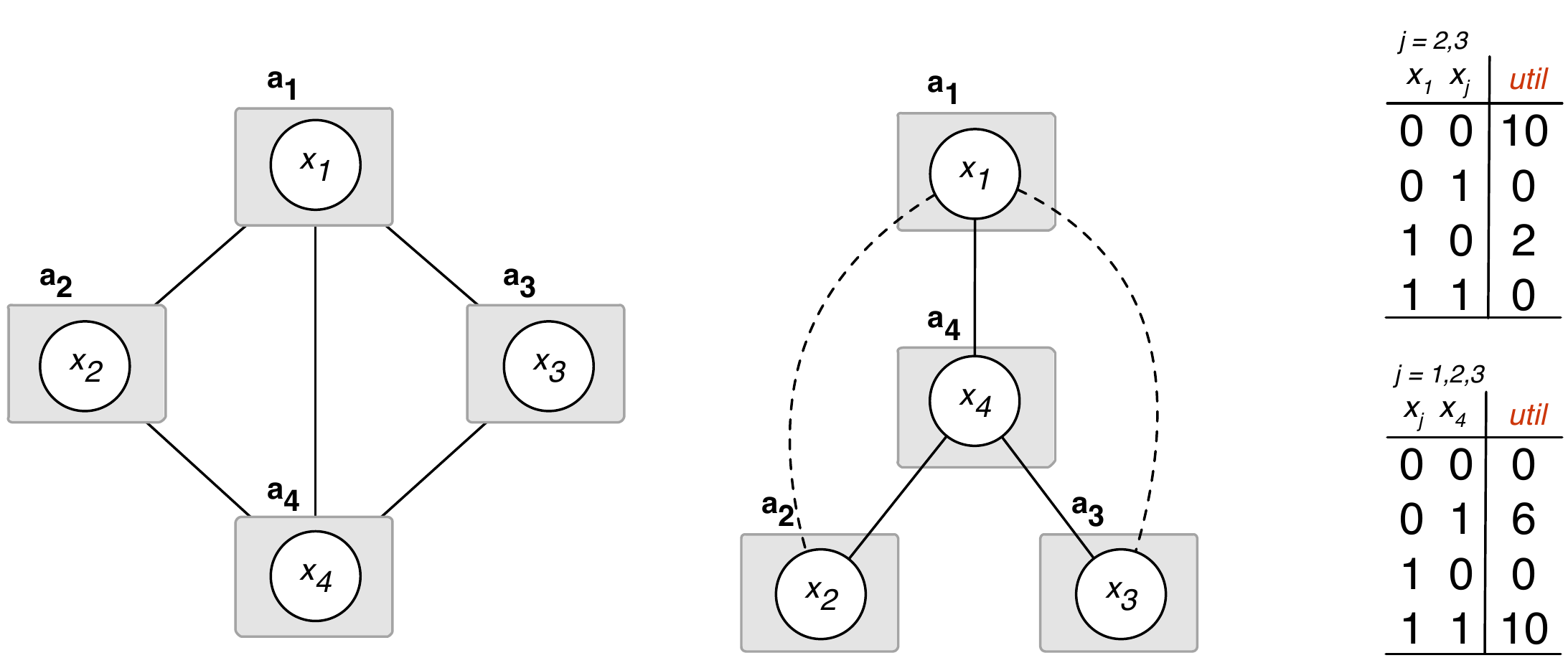}}
	  \\ \centerline{(a) Constraint Graph \hspace{0.4in} (b) Pseudo-tree \hspace{0.25in} (c) Constraints}
  \caption{Example DCOP}
\label{fig:dcop}
\end{figure}

\paragraph{Distributed Constraint Optimization Problems.}
A \emph{Distributed Constraint Optimization Problem (DCOP)} is a tuple $\langle\mathcal{X, D, F, A, \,}\alpha\rangle$, 
where:
$\mathcal{X} \!=\! \{x_1,\ldots,x_n\}$ is a set of \emph{variables}; 
$\mathcal{D} \!=\! \{D_1,\ldots,D_n\}$ is a set of finite \emph{domains} (i.e.,~$x_i \!\in\!  D_i$); 
$\mathcal{F} \!=\! \{f_1, \ldots, f_e\}$ is a set of \emph{utility functions} (also called \emph{constraints}), where $f_i: \bigtimes_{x_j \in \scope{f_i}} D_i \rightarrow \mathbb{R}_+ \cup \{-\infty\}$ and 
$\scope{f_i} \!\subseteq\! \mathcal{X}$ is the set of the variables (also called the \emph{scope}) relevant to $f_i$;
$\mathcal{A} \!=\! \{a_1, \ldots, a_p\}$ is a set of \emph{agents};
and $\alpha: \mathcal{X} \rightarrow \mathcal{A}$ is a function that maps each variable to one agent. 
$f_i$ specifies the utility of each combination of values assigned to the variables in $\scope{f_i}$.
Following common conventions, we restrict our attention to binary utility functions and  assume that  each agent controls exactly one variable. Thus, we will use the terms ``variable'' and ``agent'' interchangeably and assume that $\alpha(x_i) \is a_i$. We assume at most one constraint between each pair of variables, thus making the order of variables in the scope of a constraint irrelevant.

A \emph{partial assignment} $\sigma$ is a value assignment to a set of variables  $X_{\sigma} \!\subseteq\! \mathcal{X}$  that is consistent with the 
variables' domains. The utility $\mathcal{F}(\sigma) \!=\! \sum_{f\in{\cal F}, \scope{f} \subseteq X_{\sigma}} f(\sigma)$ 
is the sum of the utilities of all the applicable utility functions in $\sigma$. 
A \emph{solution} is a partial assignment $\sigma$ for all the variables of the problem, i.e., with $X_{\sigma} \is \mathcal X$. 
We will denote with $\setf{x}$ a solution, while $\setf{x}_i$ is the value of $x_i$ in $\setf{x}$. The goal is to find an optimal solution $\setf{x}^* = \argmax_{\setf{x}} \mathcal{F}(\setf{x})$.

Given a DCOP $P$,  $G^\ori \!=\! (\mathcal{X}, E^\ori)$ is the \emph{constraint graph} of $P$, where $(x,y) \!\in\! E^\ori$ ~iff~ $\exists f_i \!\in\! \mathcal{F}$ ~s.t.~\ $\{x,y\} \!=\! \scope{f_i}$. A \emph{DFS pseudo-tree} arrangement for $G^\ori$ is a \emph{spanning tree} {$T^\ori \!=\! \langle \mathcal{X}, E_T\rangle$} of $G^\ori$ s.t.\ if $f_i \!\in\! \mathcal{F}$ and $\{x,y\} \!\subseteq\! \scope{f_i}$, then $x$ and $y$ appear in the same branch of $T^\ori$. Edges of $G^\ori$ that are \emph{in} (resp. \emph{out} of) $E_T$ are called \emph{tree edges} (resp. \emph{backedges}). Tree edges connect a node with its parent and its children, while backedges connect a node with its \emph{pseudo-parents} and its \emph{pseudo-children}. We use $N(a_i) \!=\! \{ a_j \!\in\! \mathcal{A} \st (x_i,x_j) \!\in\! E^\ori \}$ to denote the neighbors of the agent $a_i$. 
We denote with $G^k \!=\! \langle X^k, E^k \rangle$, the subgraph of $G^\ori$ used in the execution of our iterative algorithms, where $X^k \subseteq \mathcal{X}$ and $E^k \subseteq E^\ori$.

Fig.~\ref{fig:dcop}(a) depicts the  graph of a  DCOP with  agents $a_1,\dots,a_4$, each controlling a variable with domain \{0,1\}. Fig.~\ref{fig:dcop}(b) shows a possible pseudo-tree (solid lines identify tree edges, dotted lines refer to backedges). Fig.~\ref{fig:dcop}(c) shows the DCOP constraints.

\paragraph{Large Neighborhood Search.}
In (\emph{centralized}) \emph{Large Neighborhood Search (LNS),} an initial solution is iteratively improved by repeatedly \textit{destroying} it and \textit{repairing} it. Destroying a solution means selecting a subset of variables whose current values will be discarded. The set of such variables is referred to as \textit{large neighborhood (LN).} Repairing a solution means finding a new value assignment for the destroyed variables, given that the other non-destroyed variables maintain their values from the previous iteration. 

The peculiarity of LNS, compared to other local search techniques, is  the (larger) size of the neighborhood to explore at each step. It relies on the intuition that searching over a larger neighborhood allows the process to escape local optima and find better candidate solutions.

\section{The D-LNS Framework}
\label{sec:dlns}

In this section, we introduce D-LNS, a general distributed LNS framework to solve DCOPs. Our D-LNS solutions 
need to take into account factors that are critical for the performance of distributed systems, such as network load (i.e.,~number and size of messages exchanged by agents) and the restriction that each agent is only aware of its local subproblem (i.e.,~its neighbors and the constraints whose scope includes its variables). Such properties make typical centralized LNS techniques unsuitable and infeasible for  DCOPs.

\begin{algorithm}[t]
\small{
	\caption{\algf{D-LNS}\label{alg:lns-dcop}} 

	$k \gets 0$\;
	$\langle  {\setf{x}}_i^0, \LB_i^0, \UB_i^0 \rangle \gets$ \algf{Value-Initialization}()\;
	\While{ termination condition is not met } {
		$k \gets k + 1$\;
		$z_i^k \gets$ \algf{Destroy-Algorithm}()\;
		\leIf{$z_i^k = \destroy$} {
			$\setf{x}_i^k \gets \textit{NULL}$;
		}{ $\setf{x}_i^k \gets \setf{x}_i^{k-1}$
			}
		$\langle  {\setf{x}}_i^k, \LB_i^k, \UB_i^k \rangle \gets$ 	\algf{Repair-Algorithm}($z_i^k$)\;
		\lIf{ $not$ \procf{Accept} \textnormal{(}\hspace*{0.5mm}$ {\setf{x}}_i^k,  {\setf{x}}_i^{k-1}$\textnormal{)} } {		
			$\setf{x}_i^k \gets \setf{x}_i^{k-1}$
		}
	} 
}
\end{algorithm} 

Algorithm~\ref{alg:lns-dcop} shows the general structure of D-LNS, as executed by each agent $a_i \!\in\! \mathcal A$. After initializing its iteration counter $k$ (line~1), its current value assignment $\setf{x}_i^0$ (done by randomly assigning values to variables or by exploiting domain knowledge when available), and its current lower and upper bounds $LB_i^0$ and $U\!B_i^0$ of the optimal utility (line~2), the agent, like in LNS, iterates through the destroy and repair phases (lines~3-8) until a termination condition occurs (line~3). Possible termination conditions include reaching a maximum value of $k$, a timeout limit, or a confidence threshold \mbox{on the error of the reported best solution.}

\smallskip\noindent\bemph{Destroy Phase.} 
The result of this phase is the generation of a LN, which we refer to as $LN^k \!\subseteq\! \mathcal{X}$, for each iteration $k$. This step is executed in a distributed fashion, having each agent $a_i$ calling a \textsc{Destroy-Algorithm} to determine if its local variable $x_i$ should be \bemph{destroyed} ($\destroy$) or \bemph{preserved} $(\preserve)$, as indicated by the flag $z_i^k$ (line~5). We say that destroyed (resp.~preserved) variables are (resp.~are not) in $LN^k$.
In a typical destroy process, such decisions can be either random or made by exploiting domain knowledge.
For example, in a scheduling problem, one may choose to preserve the start times of each activity and destroy the other variables. D-LNS allows the agents to use any destroy schema to achieve the desired outcome. Once the destroyed variables are determined, the agents reset their values and keep the values of the preserved variables from the previous iteration (line~6).

\smallskip\noindent\bemph{Repair Phase.} 
The agents start the repair phase, which seeks to find new value assignments for the destroyed variables, by calling a \textsc{Repair-Algorithm} (line~7). The goal of this phase is to find an improved solution by searching over a LN, which is carried exclusively by the destroyed agents. However, the step to compute the solution bounds requires the cooperation of all agents in the problem. D-LNS is general in that it does not impose any restriction on the way agents coordinate to solve this problem. We propose two distributed repair algorithms in the next section, that provide quality guarantees and online bound refinements. 
Once the agents find and evaluate a new solution, they either accept it or reject it (line~8). 
In our proposed distributed algorithms, the agents accept the solution if it does not violate any hard constraints, that is, its utility is not $-\infty$.

While most of the current incomplete DCOP algorithms fail to guarantee the consistency of the solution returned w.r.t.~the hard constraints of the problem~\cite{pearce:07
}, D-LNS can accommodate consistency checks during the repair phase. 

\section{Distributed Bounded Repair}
\label{sec:DBR}

\begin{figure}[!t]
	 \center\includegraphics[width=0.45\textwidth]{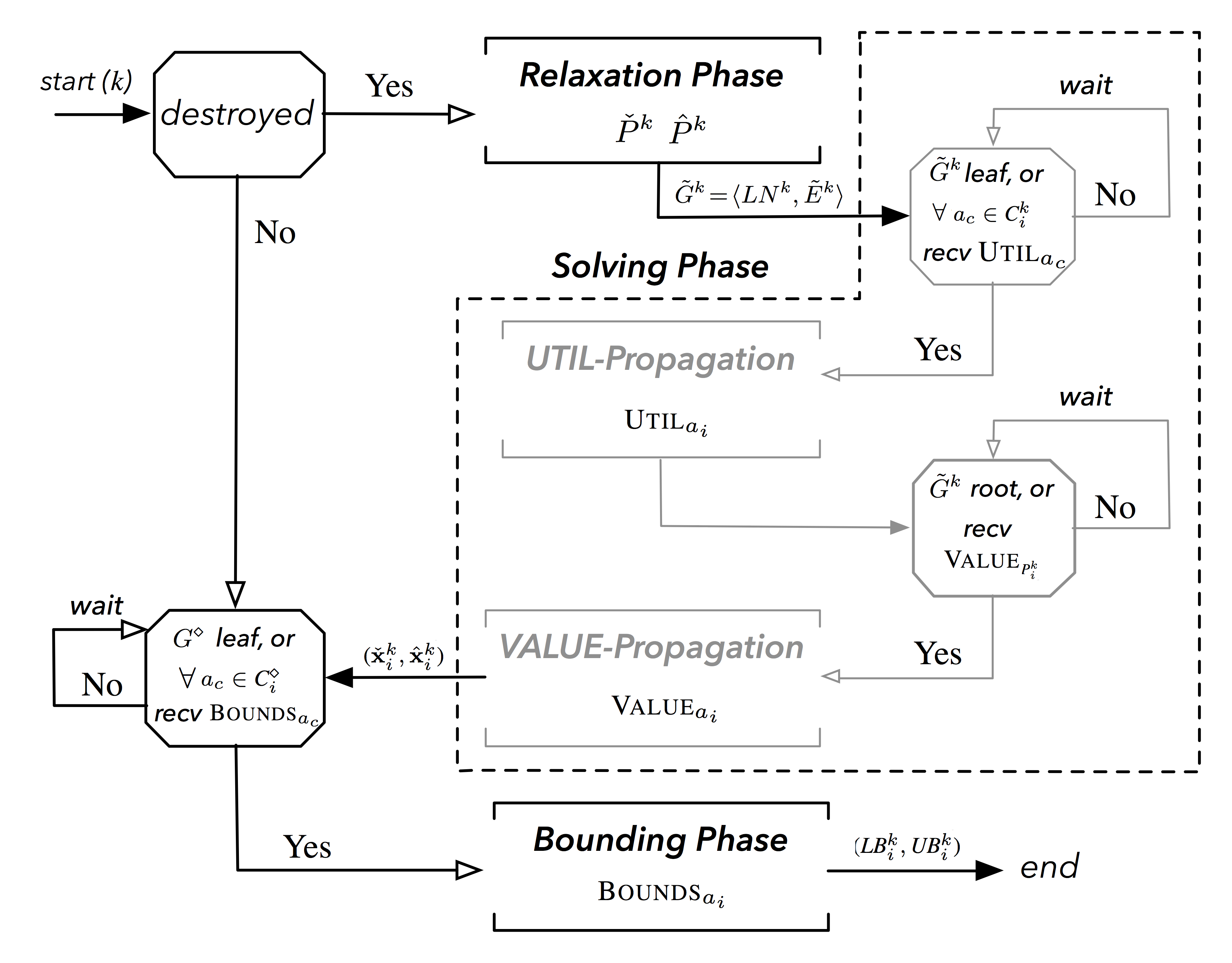}
  \caption{DBR Flow chart. The \emph{Solving phase} illustrates the T-DBR algorithm's solving phase.}
\label{fig:dbr}
\end{figure}

We now introduce the \emph{Distributed Bounded Repair} (DBR), a general \textsc{Repair} algorithm framework that, within D-LNS, iteratively refines the lower and upper bounds of the DCOP solution. 
Its general structure is illustrated in the flow chart of Figure~\ref{fig:dbr}. At each iteration $k$, each \emph{DBR} agent checks if its local variable was preserved or destroyed. In the former case, the agent waits for the \emph{Bounding} phase to start, which is algorithm dependent. In the latter case the agent executes, in order, the following phases: 

\medskip
\noindent\bemph{Relaxation Phase.}
Given a DCOP $P$, this phase constructs two \emph{relaxations} of $P$, $\check{P}^k$ and $\hat{P}^k$, which are used to compute, respectively, a lower and an upper bound on the optimal utility for $P$.
Let $G^k\!=\!\langle LN^k,E^{k} \rangle$ be the subgraph of $G^\ori$ in iteration $k$, where $E^{k} \!=\! \{(x,y) \mid (x,y) \!\in\! E^\ori; x, y \!\in\! LN^k \}$
is the subset of edges of $E^\ori$ (defined in Section 2) whose elements involve exclusively nodes in $LN^k$.
Both problem relaxations $\check{P}^k$ and $\hat{P}^k$ are solved using a \bemph{relaxation graph} $\tilde{G}^k \!=\! \langle LN^k, \tilde{E}^k \rangle$, computed from $G^k$, where $\tilde{E}^k \!\subseteq\! E^{k}$ depends on the algorithm adopted.

\smallskip\noindent
In the problem $\check{P}^k$, we wish to find a partial assignment $\underline{\check{\setf{x}}}^k$ using
\begin{equation}
	\underline{\check{\setf{x}}}^k = 
		\argmax_{\setf{x}} \Big[ 
			\sum_{f \in \tilde{E}^k} f(\setf{x}_i, \setf{x}_j)
			\hspace{6pt} + \hspace{-18pt}
			\sum_{\begin{array}{c}
				\scriptstyle{f \in {\cal F},\  \scope{f} = \{x_i,x_j\}} \\
				\scriptstyle{x_i \in LN^k,\ x_j \not\in LN^k}
				\end{array}} 
				\hspace{-26pt} f(\setf{x}_i, \check{\setf{x}}_j^{k-1}) 
				\Big]
\end{equation}
where $\check{\setf{x}}_j^{k-1}$ is the value assigned to the preserved variable $x_j$ for problem $\check{P}^{k-1}$ in the previous iteration. The first summation is over all functions listed in $\tilde{E}^k$, while the second is over all functions between destroyed and preserved variables. 
Thus, solving $\check{P}^k$ means optimizing over all the destroyed variables given that the preserved ones take on their previous value, and ignoring the (possibly empty) set of edges 
$E^\ori \setminus (\tilde{E}^k \cup \{(x,y) \mid (x,y)\!\in\!E^\ori; x\!\in\!LN^k, y\!\not\in\!LN^k \})$ 
that are not part of the relaxation graph. 
This partial assignment is used to compute lower bounds during the \emph{bounding phase}.

\smallskip\noindent
In the  problem $\hat{P}^k$, we wish to find a partial assignment $\underline{\hat{\setf{x}}}^k$ using
\begin{equation}
	\underline{\hat{\setf{x}}}^k = 
		\argmax_{\setf{x}} 
			\sum_{f \in \tilde{E}^k} f(\setf{x}_i, \setf{x}_j)
\end{equation}
Thus, solving $\hat{P}^k$ means optimizing over all the destroyed variables considering exclusively the set of edges $\tilde{E}^k$ that are part of the relaxation graph.
This partial assignment is used to compute upper bounds during the \emph{bounding phase}.

Notice that the partial assignments returned solving these two relaxed problems involve exclusively the variables in $LN^k$.

\medskip
\noindent\bemph{Solving Phase.}
Next, DBR solves the relaxed DCOPs $\check{P}^k$ and $\hat{P}^k$ using the equations above. At a high level, one can use any complete DCOP algorithm to solve $\check{P}^k$ and $\hat{P}^k$. Below, we describe two inference-based DBR algorithms, defined over different relaxation graphs $\tilde{G}^k$.
Thus, the output of this phase are the values $\check{\setf{x}}_i^k, \hat{\setf{x}}_i^k$ for the agent's local variable, associated to eqs.~(1) and (2).

\medskip
\noindent\bemph{Bounding Phase.} 
Once the relaxed problems are solved, \emph{all} agents start the bounding phase, which results in computing the lower and upper bounds based on the partial assignments $\underline{\check{\setf{x}}}^k$ and $\underline{\hat{\setf{x}}}^k$.
To do so, both solutions to the problems $\check{P}^k$ and $\hat{P}^k$ are extended to a solution $\check{\setf{x}}^k$ and $\hat{\setf{x}}^k$, respectively, for $P$, where the preserved variables $x_j \!\not\in\! LN^k$ 
are assigned the values $\check{\setf{x}}_j^{k-1}$ from the previous iteration. 

The lower bound is thus computed by evaluating  
$\mathcal{F}(\check{\setf{x}}^k)$.
The upper bound is computed by evaluating 
$\hat{F}^k(\hat{\setf{x}}^k) = \sum_{f \in \mathcal{F}} \hat{f}^k(\hat{\setf{x}}^k_i, \hat{\setf{x}}^k_j)$,
where 
%
\begin{align} 
\!\!\!\hat{f}^k(\setf{x}_i, \setf{x}_j) 
\is \begin{cases}
	\displaystyle 
	\max_{d_i \in D_i, d_j \in D_j} \hspace{-4pt} f(d_i, d_j)  
		& \text{if } \Gamma_f^k = \emptyset\\
	\max \Big\{ 
			\frac{\tilde{F}^k}{ |\tilde{E}^k| }, 
			\displaystyle\max_{{\scriptscriptstyle \ell \in \Gamma_f^{k-1}}} \hspace{-2pt} 
			\hat{f}^{\ell}(\hat{\setf{x}}_i^{\ell}, \hat{\setf{x}}_j^{\ell}) \Big \} \hspace{-6pt} 
		& \text{if } f \in \tilde{E}^k\\
	\hat{f}^{k-1}(\hat{\setf{x}}_i^{k-1}, \hat{\setf{x}}_j^{k-1})
		& \text{otherwise} \\
\end{cases}
\label{eq:fhat}
\end{align}
with $\tilde{F}^k = \max_{\setf{x}} \sum_{f \in \tilde{E}^k} f(\setf{x}_i, \setf{x}_j)$ is the optimal utility on the relaxation graph $\tilde{G}^k$, and 
$\Gamma_f^k$ is the set of past iteration indices for which the function $f$ was an edge in the relaxation graph. Specifically,
$
\Gamma_f^k = \left\{ \ell \st f \in \tilde{E}^{\ell}
\land 0 < \ell \leq k \right\}
$. 

Therefore, the utility of $\hat{F}^k(\hat{\setf{x}}^k)$ is composed of three parts.
The first part involves all functions that have never been part of $\tilde{E}^k$ up to the current iteration,
the second part involves all the functions optimized in the current iteration, and
the third part involves all the remaining functions.
The utility of each function in the first part is the maximal utility over all possible pairs of value combinations of variables in the scope of that function.
The utility of each function in the second part is the largest utility among the mean utility of the functions optimized in the current iteration (i.e., those in $\tilde{E}^k$), and the utilities of such function optimized in a past iteration. 
The utility of each function in the third part is equal to the utility assigned to such function in the previous iteration.
In particular, imposing that the edges optimized in the current iteration contribute at most equally (i.e., as the mean utility of $\tilde{F}^k$) to the final utility of $\hat{P}^k$ 
allows us to not underestimate the solution upper bound within the iterative process (see Lemma~1).
%
As we show in Theorems~\ref{th:lb} and \ref{th:ub}, \mbox{$\mathcal{F}(\check{\setf{x}}^k)\!\leq\! \mathcal{F}(\setf{x}^*)\!\leq$} $\!\hat{F}^k(\setf{\hat{x}}_k)$. Therefore, 
$
	\rho \!=\! \frac{ \min_{k} \hat{F}^k(\hat{\setf{x}}^k) }
						 { \max_k \mathcal{F}(\check{\setf{x}}^k) }
$
is a guaranteed approximation ratio for $P$. 

\smallskip 
The significance of this \textsc{Repair} framework is that it enables D-LNS to iteratively refine both lower and upper bounds of the solution, without imposing any restrictions on the form of the objective function and of the constraints adopted.\footnote{
Note, however that this does not implies that the lower bound and the upper bound will converge to the same value.} 
Below, we introduce two implementations of the DBR framework, summarized in the flow-chart of Figure \ref{fig:dbr}, whose solving phase is shown in the dotted area.

\subsection{DPOP-based DBR Algorithm}
DPOP-based DBR (DPOP-DBR) solves the relaxed DCOPs $\check{P}^k$ and $\hat{P}^k$ over the relaxed graph $\tilde{G}^k \!=\! \langle LN^k, E^k\rangle$. Thus, $\tilde{E}^k \!=\! E^k$, and solving problem $\check{P}^k$ means optimizing over all the destroyed variables ignoring no edges in $E^k$.

The DPOP-DBR \emph{solving phase} uses DPOP~\cite{petcu:05}, a complete inference-based algorithm  composed of two phases operating on a DFS pseudo-tree. In the \bemph{utility propagation phase}, each agent, starting from the leaves of the pseudo-tree, projects out its own variable and sends its projected utilities to its parent. These utilities are propagated up the pseudo-tree induced from $\tilde{G}^k$ until they reach the root. The hard constraints of the problem can be naturally handled in this phase, by  pruning all inconsistent values before sending a message to its parent. Once the root receives utilities from all its children, it starts the \bemph{value propagation phase}, where it selects the value that maximizes its utility and sends it  to its children, which repeat the same process. The problem is solved as soon as  the values  reach  the leaves.

Note that the relaxation process may create a forest, in which case one should execute the algorithm in each tree of the forest. As a technical note, DPOP-DBR solves the two relaxed DCOPs in parallel. In the utility propagation, each agent computes two sets of utilities, one for each relaxed problem, and sends them to its parent. In the value propagation phase, each agent selects two values, one for each relaxed problem, and sends them to its children.

DPOP-DBR has the same worst case order complexity of DPOP, that is, exponential in the induced width of the relaxed graph $\tilde{G}^k$. Thus, we introduce another algorithm characterized by a smaller complexity and low network load.

\subsection{Tree-based DBR Algorithm}
\emph{Tree-based DBR (T-DBR)} defines the relaxed DCOPs $\check{P}^k$ and $\hat{P}^k$ using a pseudo-tree structure {$T^k \!=\! \langle LN^k, E_{T^k} \rangle$} that is computed from the subgraph $G^k$. 
Thus, $\tilde{E}^k \!=\! E_{T^k}$, and solving problem $\check{P}^k$ means optimizing over all the destroyed variables ignoring backedges. 
Its general solving schema is similar to that of DPOP, in that it uses Utility and Value propagation phases; however, the different underlying relaxation graph adopted imposes several important differences. 
Algorithm~\ref{alg:blns} shows the T-DBR pseudocode.
%
We use the following notations: $P_i^k$, $C_i^k$, 
$P\!P_i^k$ denote the parent, the set of children, 
and pseudo-parents of the agent $a_i$, at iteration $k$. The set of these items is referred to as $\setf{T}_i^k$, which is  $a_i$'s \emph{local view} of the pseudo-tree $T^k$.  We use ``$^\ori$'' to refer to the items associated with the pseudo-tree $T^\ori$. $\check{\chi}_i$ and $\hat{\chi}_i$ denote $a_i$'s \emph{context} (i.e., the values for each $x_j \in N(a_i)$) w.r.t.~problems $\check{P}$ and $\hat{P}$, respectively. We assume that by the end of the destroy phase (line~6) each agent knows its current context as well as which of its neighboring agents has been destroyed or preserved.
In each iteration $k$, T-DBR executes the following phases:

\SetAlgoRefName{2}

\begin{algorithm}[!t]
\small{%
	\caption{ \algf{T-DBR}($z_i^k$)} \label{alg:blns}
	
	$\setf{T}_i^k \gets $ \algf{Relaxation}($z_i^k$)\\
	\algf{Util-Propagation}($\setf{T}_i^k$)\\
	$\langle \check{\chi}_i^k, \hat{\chi}_i^k \rangle \gets$ \algf{Value-Propagation}($\setf{T}_i^k$)\\
	$\langle \LB_i^k, \UB_i^k \rangle \gets$ \algf{Bound-Propagation}($\check{\chi}_i^k, \hat{\chi}_i^k$)\\
	\Return{$\langle \check{\setf{x}}_i^k, \LB_i^k, \UB_i^k \rangle$}\\
}%
\end{algorithm}
%
%
\begin{procedure}[!t]
\small{%
	\caption{{UTIL}-Propagation($\mathbf{T}_i^k$)}\label{alg:blns-util}
	
	\textbf{receive} \msgf{Util}$_{a_c}(\varf{\check{U}_c, \hat{U}_c})$ from each $a_c \in \textit{C}_i^k$\\

	\ForAll{values $\setf{x}_i, \setf{x}_{\textit{P}_i^k}$}{
	$\varf{\check{U}}_i(\setf{x}_i, \setf{x}_{\textit{P}_i^k}) \gets f(\setf{x}_i, \setf{x}_{\textit{P}_i^k}) + $ 
	$~\hspace{12pt} \sum_{a_c \in \textit{C}_i^k}{
	 \varf{\check{U}}_c}(\setf{x}_i) 
	 \hspace{0pt} + 
 	\sum_{x_j \not\in \textit{LN}^k} 
	f(\setf{x}_i, \check{\setf{x}}_j^{k-1})$\\

	$\varf{\hat{U}}_i(\setf{x}_i, \setf{x}_{\textit{P}_i^k}) \gets 
	 f(\setf{x}_i, \setf{x}_{\textit{P}_i^k}) 
	 \hspace{0pt} + 
	 \sum_{a_c \in \textit{C}_i^k}{
	 \varf{\hat{U}}_c}(\setf{x}_i)$\\

	}
	\ForAll{values $\setf{x}_{\textit{P}_i^k}$}{
	$
	\hspace{-4pt}
	\langle \varf{\check{U}}_i'(\setf{x}_{\textit{P}_i^k}), \varf{\hat{U}}_i'(\setf{x}_{\textit{P}_i^k}) \rangle
	\hspace{-2pt}
	\gets
	\hspace{-2pt}
	\displaystyle 
	\langle \max_{\setf{x}_i} \varf{\check{U}}_i(\! \setf{x}_i \!, \setf{x}_{\textit{P}_i^k} \!)\!, 
	            \max_{\setf{x}_i} \varf{\hat{U}}_i(\! \setf{x}_i \!, \setf{x}_{\textit{P}_i^k} \!) 
	\rangle
	$
	}
	\textbf{send} \msgf{Util}$_{a_i}(\varf{\check{U}_i', \hat{U}_i'})$ msg to $\textit{P}_i^k$\\
}
\end{procedure} 
%
\begin{function}[!t]
\small{%
	\caption{VALUE-Propagation($\mathbf{T}_i^k$)}
	\eIf{$\textit{P}_i^k =$ \textit{NULL}}{	
		$\langle \check{\setf{x}}_i^k, \hat{\setf{x}}_i^k \rangle 
		\gets 
		\langle 
			\argmax_{\setf{x}_i} \check{U}_i(\setf{x}_i), 
			\argmax_{\setf{x}_i} \hat{U}_i(\setf{x}_i) 
		\rangle$\\
		\textbf{send} \msgf{Value}$_{a_i}$($\check{\setf{x}}_i^k, \hat{\setf{x}}_i^k$) msg to each $a_j \in N(a_i)$\\
		\ForAll{$a_j \in N(a_i)$}{
			\textbf{receive} \msgf{Value$_{a_j}$}($\check{\setf{x}}_{j}^k, \hat{\setf{x}}_{j}^k$) msg from $a_j$\\
			Update $x_j$ in 
			$\langle \check{\chi}_i^k, \hat{\chi}_i^k \rangle$ 
			with $\langle \check{\setf{x}}_j^k$, $\hat{\setf{x}}_j^k \rangle$\\			
		}
	}
	{
		\ForAll{$a_j \in N(a_i)$}{
			\textbf{receive} \msgf{Value$_{a_j}$}($\check{\setf{x}}_{j}^k, \hat{\setf{x}}_{j}^k$) msg from $a_j$\\
			Update $x_j$ in 
			$\langle \check{\chi}_i^k, \hat{\chi}_i^k \rangle$ 
			with $\langle \check{\setf{x}}_j^k$, $\hat{\setf{x}}_j^k \rangle$\\
			\If{$a_j = \textit{P}_i^k$}{
				$
				\hspace{-2pt}
				\langle \check{\setf{x}}_i^k, \hat{\setf{x}}_i^k \rangle 
				\hspace{-2pt}
				\gets
				\hspace{-2pt}
				\langle
				\argmax_{\setf{x}_i} \hspace{-3pt} \check{U}_i(\setf{x}_i),
				\argmax_{\setf{x}_i} \hspace{-3pt} \hat{U}_i(\setf{x}_i)
				\rangle$\\
				\hspace{-1pt}\textbf{send} \msgf{Value}$_{a_i}$($\check{\setf{x}}_i^k, \hat{\setf{x}}_i^k$) msg to each $a_j\! \in \!N(a_i)$\\

			}
		}
	}
	\Return{$\langle \check{\chi}_i^k, \hat{\chi}_i^k  \rangle$}
}
\end{function} 
%
\begin{procedure}[!t]
\small{%
	\caption{BOUND-Propagation($\check{\chi}_i^k, \hat{\chi}_i^k $)}
	\label{alg:blns-bounds}

	\textbf{receive} \msgf{Bounds$_{a_c}$}($\LB_c^k, \UB_c^k$) msg from each $a_c \in \textit{C}_i^\ori$\\

	$\LB_i^k 
	\hspace{-2pt}\gets\hspace{-2pt}
	 f(\check{\setf{x}}_i^k, \check{\setf{x}}_{\textit{P}_i^\ori}^k) + \hspace{-0pt}
	 \sum_{a_j \in \textit{PP}_i^\ori} \hspace{-0pt} f(\check{\setf{x}}_i^k, \check{\setf{x}}_j^k) + \hspace{-0pt}
 	 \sum_{a_c \in \textit{C}_i^\ori} \LB_c^k$\\

	$\textit{UB}_i^k 
	\hspace{-2pt}\gets\hspace{-2pt}
	 \hat{f}^k(\hat{\setf{x}}_i, \hat{\setf{x}}_{\textit{P}_i^\ori}) + \hspace{-0pt}
	 \sum_{a_j \in \textit{PP}_i^\ori} \hspace{-0pt} \hat{f}^k(\hat{\setf{x}}_i, \hat{\setf{x}}_j) + \hspace{-0pt}
	 \sum_{a_c \in \textit{C}_i^\ori} \textit{UB}_c^k$\\

	\textbf{send} \msgf{Bounds$_{a_i}$}($\LB_i^k, \UB_i^k$) msg to $\textit{P}_i^\ori$\\

}
\end{procedure}

%
%

\smallskip
\noindent\bemph{Relaxation Phase.} 
It constructs a pseudo-tree $T^k$ (line~9), which ignores, from $G^\ori$, the preserved variables as well as the functions involving these variables in their scopes. The construction prioritizes tree-edges that have not been chosen in previous pseudo-trees over the others. 

\begin{figure*}[t]
	\centering\includegraphics[width=0.92\textwidth]{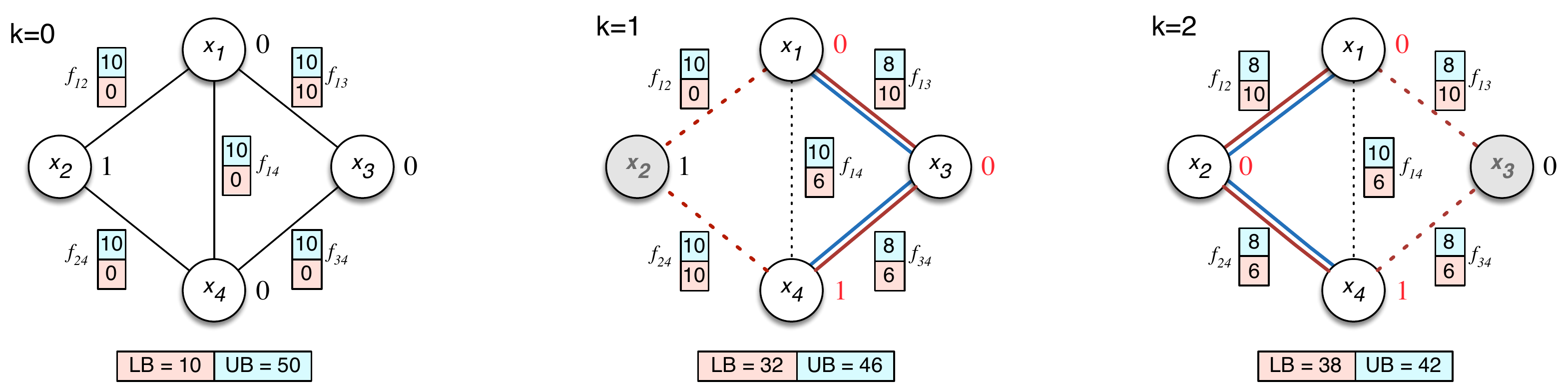} 
  \caption{D-LNS with T-DBR example trace.
  \label{fig:blns}}
\end{figure*}

\smallskip
\noindent\bemph{Solving Phase.} 
Similarly to DPOP-DBR, T-DBR solving phase is composed of two phases operating on the relaxed pseudo-tree $T^k$, and executed synchronously:
\bitemize
\item \emph{Utility Propagation Phase.} 
After the pseudo-tree $T^k$ is constructed (line~10), each leaf agent computes the optimal sum of utilities in its subtree considering exclusively tree edges (i.e., edges in $E_{T^k}$) and edges with destroyed variables. Each leaf agent computes the utilities 
$\check{U}_i(\setf{x}_i, \setf{x}_{\textit{P}_i^k})$ and $\hat{U}_i(\setf{x}_i, \setf{x}_{\textit{P}_i^k})$ for each pair of values of its variable $\setf{x}_i$ and its parent's variable $\setf{x}_{\textit{P}_i^k}$ (lines~15-17), in preparation for retrieving the solutions of 
$\check{P}$ and $\hat{P}$, used during the bounding phase.
The agent projects  itself out (lines~18-19) and sends the projected utilities  to its parent in a \textsc{Util} message (line~20). Each agent, upon receiving the \textsc{Util} message from each child, performs the same operations. Thus, these utilities will propagate up the pseudo-tree until they reach the root agent.	
	
\item \emph{Value Propagation Phase.} 
This phase starts after the utility propagation (line 11) by having the root agent compute its optimal values $\check{\setf{x}}_i^k$ and $\hat{\setf{x}}_i^k$ for the relaxed DCOPs $\check{P}$ and $\hat{P}$, respectively (line~22). It then sends its values to all its neighbors in a \textsc{Value} message (line~23). When its child receives this message, it also compute its optimal values and sends them to all its neighbors (lines~31-33). Thus, these values propagate down the pseudo-tree until they reach the leaves, at which point every agent has chosen its respective values. In this phase, in preparation for the bounding phase, when each agent receives a \textsc{Value} message from its neighbor, it will also update the value of its neighbor in both its contexts $\check{\chi}_i^k$ and $\hat{\chi}_i^k$ (lines~24-26 and 29-30). 
\eitemize

\smallskip
\noindent\bemph{Bounding Phase.} 
{Once the relaxed DCOPs $\check{P}$ and $\hat{P}$ have been solved}, the algorithm starts the bound propagation phase (line~12). This phase starts by having each leaf agent of the pseudo-tree $T^\ori$ compute the lower and upper bounds $\LB_i^k$ and $\UB_i^k$ (lines~36-37). These bounds are sent to its parent in $T^\ori$ (line~38). When its parent receives this message form all its children (line~35), it performs the same operations. The lower and upper bounds of the whole problem are determined
when the  bounds reach  the root agent.

\subsection{T-DBR Example Trace}
Figure~\ref{fig:blns} illustrates a running example of T-DBR during the first two D-LNS iterations, using the DCOP of Figure~\ref{fig:dcop}. 
The trees $T^1$ and $T^2$ are represented by bold solid lines (functions in $E_{T^k}$);  all other functions are represented by dotted lines. 
The preserved variables in each iteration are shaded gray. 
At each step, the resolution of the relaxed problems involves the functions represented by bold lines---$\hat{P}$ is solved optimizing over the blue colored functions, and $\check{P}$ over the red ones.
We recall that while solving $\hat{P}$ focuses solely on the functions in $E_{T^k}$, 
solving $\check{P}$ further accounts for the function involving a destroyed and a preserved variable.
The nodes illustrating destroyed variables are labeled with red values representing $\check{\setf{x}}_i^k$,\footnote{In our example solving $\check{P}$ and $\hat{P}$ yields the same solution for $k=1,2$.} and nodes representing preserved variables are labeled with black values representing $\check{\setf{x}}_i^{k-1}$. 
Each edge is labeled with a pair of values representing the utilities $\hat{f}^k(\check{\setf{x}}_i^k, \check{\setf{x}}_j^k)$ (top, in blue) and $f(\check{\setf{x}}_i^k, \check{\setf{x}}_j^k)$ (bottom, in red) of the corresponding functions. The lower and upper bounds of each iteration are shown below. 

When $k \is 0$, each agent randomly assigns a value to its variable, which results in a solution with utility 
$\mathcal{F}(\check{\setf{x}}^0) \is 
f(\check{\setf{x}}_1^0, \check{\setf{x}}_2^0) \!+\!
f(\check{\setf{x}}_1^0, \check{\setf{x}}_3^0) \!+\!
f(\check{\setf{x}}_1^0, \check{\setf{x}}_4^0) \!+\!
f(\check{\setf{x}}_2^0, \check{\setf{x}}_4^0) \!+\!
f(\check{\setf{x}}_3^0, \check{\setf{x}}_4^0) \is  0 \!+\! 10 \!+\! 0 \!+\! 0 \!+\! 0 \is  10$ to get the lower bound. 
Moreover, solving $\hat{P}^0$ yields a solution $\hat{\setf{x}}^0$ with utility 
$\hat{F}^0(\hat{\setf{x}}^0)  \is  
\hat{f}^0(\hat{\setf{x}}_1^0, \hat{\setf{x}}_2^0) \!+\!
\hat{f}^0(\hat{\setf{x}}_1^0, \hat{\setf{x}}_3^0) \!+\!
\hat{f}^0(\hat{\setf{x}}_1^0, \hat{\setf{x}}_4^0) \!+\!
\hat{f}^0(\hat{\setf{x}}_2^0, \hat{\setf{x}}_4^0) \!+\!
\hat{f}^0(\hat{\setf{x}}_3^0, \hat{\setf{x}}_4^0)  \is  10 \!+\! 10 \!+\! 10 \!+\! 10 \!+\! 10  \is  50$, which is the upper bound. 

In the first iteration ($k \is 1$), the destroy phase preserves $x_2$, and thus $\check{\setf{x}}_2^1  \is  \check{\setf{x}}_2^0  \is  1$. 
The algorithm then builds the spanning tree with the remaining variables choosing 
$f_{13}$ and $f_{34}$ as a tree edges. Thus the relaxation graph for $\check{P}^1$ involves the edges $\{f_{13}, f_{34}, f_{12}, f_{24}\}$ (in red), and the relaxation graph for $\hat{P}^1$ involves the edges $\{f_{13}, f_{34}\}$ (in blue). 
Solving $\check{P}^1$ yields partial assignment $\check{\setf{\underline{x}}}^1$ with utility $\check{F}^1(\check{\setf{\underline{x}}}^1)  \is  
f(\check{\setf{x}}_1^1, \check{\setf{x}}_3^1) \!+\!
f(\check{\setf{x}}_3^1, \check{\setf{x}}_4^1) \!+\!
f(\check{\setf{x}}_1^1, \check{\setf{x}}_2^1) \!+\!
f(\check{\setf{x}}_2^1, \check{\setf{x}}_4^1)  \is  10 + 6 + 0 + 10  \is  26$, which results in a lower bound 
$\mathcal{F}(\check{\setf{x}}^1)  \is  \check{F}^1(\check{\setf{\underline{x}}}^1) + f(\check{\setf{x}}_1^1, \check{\setf{x}}_4^1)  \is  26 + 6  \is  32$. 
Solving $\hat{P}^1$ yields solution $\hat{\setf{x}}^1$ with utility 
$\hat{F}^1(\hat{\setf{x}}^1)  \is  
\hat{f}^1(\hat{\setf{x}}_1^1, \hat{\setf{x}}_2^1) \!+\!
\hat{f}^1(\hat{\setf{x}}_1^1, \hat{\setf{x}}_3^1) \!+\!
\hat{f}^1(\hat{\setf{x}}_1^1, \hat{\setf{x}}_4^1) \!+\!
\hat{f}^1(\hat{\setf{x}}_2^1, \hat{\setf{x}}_4^1) \!+\!
\hat{f}^1(\hat{\setf{x}}_3^1, \hat{\setf{x}}_4^1)  \is  10 \!+\! 8 \!+\! 10 \!+\! 10 \!+\! 8  \is  46$, which is the current upper bound. 
Recall that the values for the functions in $\tilde{E}^k$ are computed as  $\frac{\tilde{F}^k(\setf{x})}{ |\tilde{E}^k| } \!=\! \frac{16}{2} \!=\! 8$ (see eq.~ \eqref{eq:fhat}).

Finally, in the second iteration ($k \is 2$), the destroy phase retains $x_3$ assigning it its value in the previous iteration $\check{\setf{x}}_3^2  \is  \check{\setf{x}}_3^1  \is  0$, and the repair phase builds the new spanning tree with the remaining variables choosing 
$f_{12}$ and $f_{24}$ as a tree edges. Thus the relaxation graph for $\check{P}^2$ involves the edges $\{f_{12}, f_{24}, f_{13}, f_{34}\}$, and the relaxation graph for $\hat{P}^2$ involves the edges $\{f_{12}, f_{24}\}$. 
Solving $\check{P}^2$ and $\hat{P}^2$ yields partial assignments $\check{\setf{\underline{x}}}^2$ and $\hat{\setf{\underline{x}}}^2$, respectively, with utilities  $\check{F}^2(\check{\setf{\underline{x}}}^2)  \is  10 \!+\! 6 \!+\! 10 \!+\! 6  \is  32$, which results in a lower bound $\mathcal{F}(\check{\setf{x}}^2)  \is  32 \!+\! 6  \is  38$, and an upper bound $\hat{F}^2(\hat{\setf{x}}^2)  \is  8 \!+\! 8 \!+\! 10 \!+\! 8 \!+\! 8  \is  42$.

\section{Theoretical Properties}
\label{sec:theo}
We report below the theoretical results on the bounds provided by our D-LNS framework with the DBR \textsc{Repair} algorithm, as well as the agents' complexity and network load of T-DBR. Due to space constraints, we report sketch proofs.

\noindent
\begin{theorem}
	\label{th:lb} 
	For each  $LN^k$, 
	$
		\mathcal{F}(\check{\setf{x}}^k) \leq \mathcal{F}(\setf{x}^*).
	$
\end{theorem}
\begin{sproof}
The result follows from that $\check{\setf{\underline{x}}}^k$ is an optimal solution of the relaxed problem $\check{P}$ whose functions are a subset of $\mathcal{F}$.
\end{sproof}

\begin{lemma}
	\label{pr:ub}
For each $k$,
	$
	\displaystyle
	\sum_{f \in \tilde{E}^k} \hat{f}(\hat{\setf{x}}_i^k, \hat{\setf{x}}_j^k) 
	\geq 
	\sum_{f \in \tilde{E}^k} f(\setf{x}_i^*, \setf{x}_j^*),
	$
	where $\hat{\setf{x}}_i^k$ is the value assignment to variable $x_i$ when solving the relaxed DCOP $\hat{P}$ and $\setf{x}_i^*$ is the value assignment to variable $x_i$ when solving the original DCOP $P$.
\end{lemma}
\begin{sproof}
For each iteration $k$, it follows:
\begin{align*}
	\sum_{f \in \tilde{E}^k} \hspace{-2pt} \hat{f}(\hat{\setf{x}}_i^k, \hat{\setf{x}}_j^k) 
	&\geq \hspace{-2pt}
	\sum_{f \in \tilde{E}^k} \hspace{-3pt}
		\max \Big\{ 
			\frac{\tilde{F}^k}{ |\tilde{E}^k| }, 
			\displaystyle\max_{{\scriptscriptstyle \ell \in \Gamma_f^{k-1}}} \hspace{-2pt} 
			\hat{f}^{\ell}(\hat{\setf{x}}_i^{\ell}, \hat{\setf{x}}_j^{\ell}) \Big \}
	\tag{by def.~of $\hat{f}$ (case 2)} \\
	&\geq 
	\sum_{f \in \tilde{E}^k} \frac{\tilde{F^k}}{|\tilde{E}^k|} \\
	\geq 
	\tilde{F^k}
	&= 
	\sum_{f \in \tilde{E}^k} f(\hat{\setf{x}}_i^k, \hat{\setf{x}}_j^k) 
	\tag{by def.~of $\tilde{F}^k$}\\
	&\geq
	\sum_{f \in \tilde{E}^k} f(\hat{\setf{x}}_i^*, \hat{\setf{x}}_j^*).
\end{align*}
The last step follows from that, in each iteration $k$, the functions associated with the edges in $\tilde{E}^k$ are solved optimally. Since their cost is maximized it is also greater than the corresponding cost when evaluated on the optimal solution for the problem $P$. 
\end{sproof}


\begin{lemma}
	\label{lm:ub}
For each $k$, 
	$
	\displaystyle \sum_{f \in \Theta^k} \hat{f}^k(\setf{\hat{x}}_i^k,\setf{\hat{x}}_j^k) \geq 
		\sum_{f \in \Theta^k} f(\setf{x}_i^*, \setf{x}_j^*),
	$
where $\Theta^k = \{ f \in \mathcal{F} \st \Gamma^k_f \neq \emptyset\}$ is the set of functions that have been chosen as edges of the relaxation graph in a previous iteration.	
\end{lemma}
\begin{sproof}
We prove it by induction on the iteration $k$.
For ease of explanation we provide an illustration (below) of the set of relevant edges optimized in successive iterations. 
\begin{figure}[!h]
\center\includegraphics[width=0.27\textwidth]{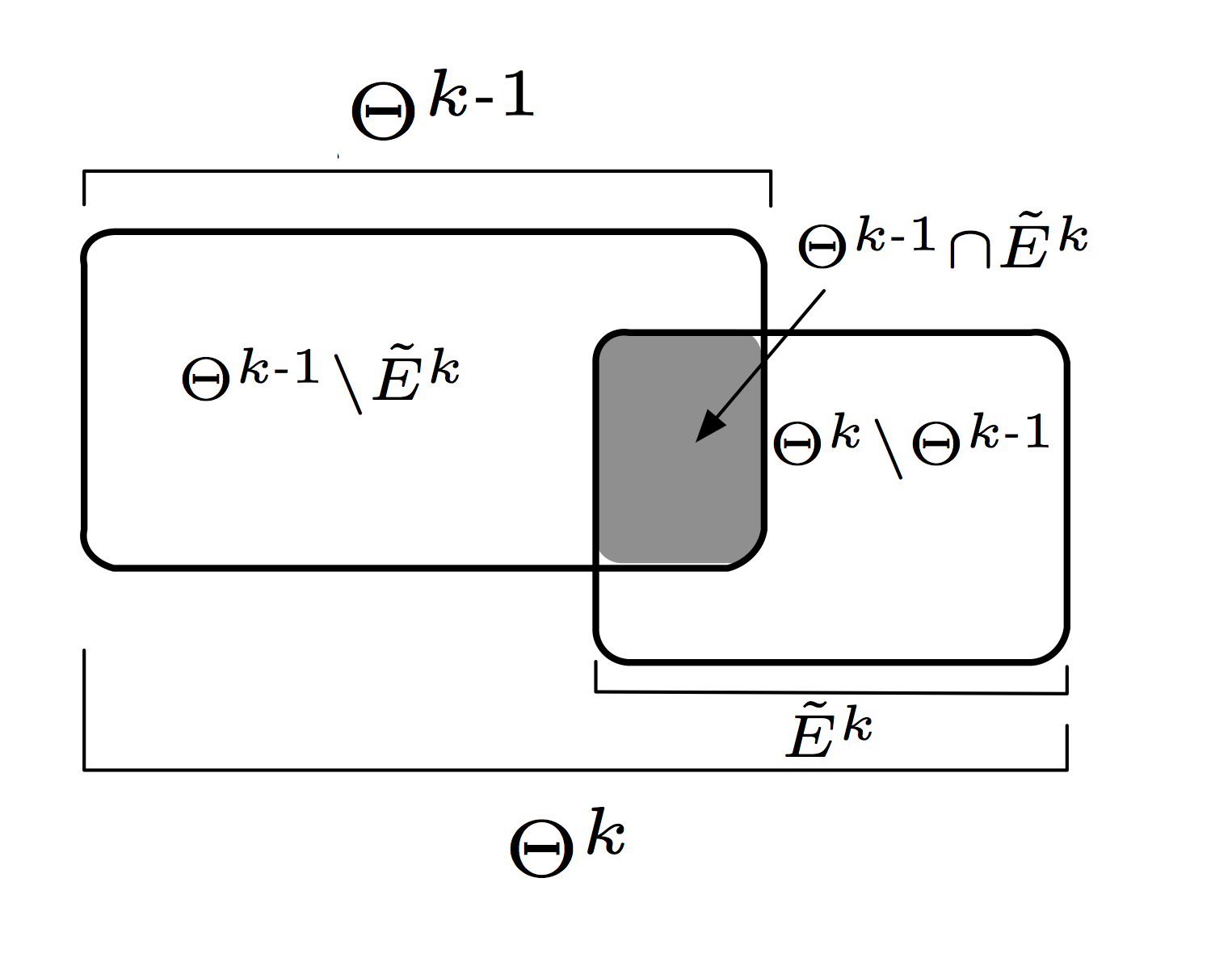}
\label{fig:proof:1}
\end{figure}

\noindent
For $k \!=\! 0$, then $\Theta^0 \!=\! \emptyset$, thus the statement vacuously holds. Assume the claim holds up to iteration $k-1$. For iteration $k$ it follows that,
%
\begin{align*}
\hspace{10pt} 
	& \hspace{-10pt} \sum_{f \in \Theta^{k}}\! \hat{f}^{k}(\setf{\hat{x}}_i^k,\setf{\hat{x}}_j^k) \notag \\
      &= \hspace{-4pt}  \sum_{f \in \Theta^{k\minus1}} \hspace{-4pt} \hat{f}^{k}(\setf{\hat{x}}_i^k,\setf{\hat{x}}_j^k) 
	+ 
	\hspace{-4pt} \sum_{f \in \Theta^{k} \setminus \Theta^{k\minus1}} \hspace{-12pt} \hat{f}^{k}(\setf{\hat{x}}_i^k,\setf{\hat{x}}_j^k) \\
	\hspace{12pt}
	&= 
	\hspace{-12pt} 
	\sum_{f \in \Theta^{k\minus1} \setminus \tilde{E}^k} \hspace{-12pt} \hat{f}^{k}(\setf{\hat{x}}_i^k,\setf{\hat{x}}_j^k)
	+ \hspace{-12pt}
	 \sum_{f \in \Theta^{k\minus1} \cap \tilde{E}^k} \hspace{-12pt} \hat{f}^{k}(\setf{\hat{x}}_i^k,\setf{\hat{x}}_j^k) 
	+ \hspace{-12pt}
	\sum_{f \in \Theta^{k} \setminus \Theta^{k\minus1}} \hspace{-12pt} \hat{f}^{k}(\setf{\hat{x}}_i^k,\setf{\hat{x}}_j^k)  \\
	&= \hspace{-12pt}
	\sum_{f \in \Theta^{k\minus1} \setminus \tilde{E}^{k}} \hspace{-12pt} \hat{f}^{k\minus1}(\setf{\hat{x}}_i^{k\minus1},\setf{\hat{x}}_j^{k\minus1})
	+ \hspace{-12pt} 	\sum_{f \in \Theta^{k} \setminus \Theta^{k\minus1}} \hspace{-12pt} \hat{f}^{k}(\setf{\hat{x}}_i^k,\setf{\hat{x}}_j^k) \notag \\
	&\hspace{12pt} + \hspace{-12pt} 
	\sum_{f \in \Theta^{k\minus1} \cap \tilde{E}^k} \hspace{-12pt} 
		\max \big\{ \hat{f}^k(\hat{\setf{x}}_{i~,}^{k} \hat{\setf{x}}_j^{k}), 
					 \hat{f}^{k\minus1}(\setf{\hat{x}}_{i ~~,}^{k\minus1} \setf{\hat{x}}_j^{k\minus1}) \big\} 
			\tag{by def.~of $\hat{f}^{k}$} 
\end{align*}
The last step follows from cases 2 and 3 of eq.~\eqref{eq:fhat}.
Additionally, the following inequalities hold:
\begin{align*}
& \hspace{-12pt}
	\sum_{f \in \Theta^{k\minus1} \setminus \tilde{E}^{k}} \hspace{-12pt} \hat{f}^{k\minus1}(\setf{\hat{x}}_i^{k\minus1},\setf{\hat{x}}_j^{k\minus1}) \notag 
	+ \hspace{-12pt} 
	\sum_{f \in \Theta^{k\minus1} \cap \tilde{E}^k} \hspace{-12pt} 
		\max \big\{ \hat{f}^k(\hat{\setf{x}}_{i~,}^{k} \hat{\setf{x}}_j^{k}), 
					 \hat{f}^{k\minus1}(\setf{\hat{x}}_{i ~~,}^{k\minus1} \setf{\hat{x}}_j^{k\minus1}) \big\}  \notag \\
	\hspace{12pt}
	&\geq 
	\sum_{f \in \Theta^{k\minus1}} \hat{f}^{k\minus1}(\setf{\hat{x}}_i^{k\minus1},\setf{\hat{x}}_j^{k\minus1}) \notag \\
	\hspace{12pt}
	&\geq 
	\sum_{f \in \Theta^{k\minus1}} f(\setf{x}_i^*,\setf{x}_j^*). \tag{by inductive hypothesis}\\
& \hspace{-12pt}
	\sum_{f \in \Theta^{k\minus1} \cap \tilde{E}^k} \hspace{-12pt} 
		\max \big\{ \hat{f}^k(\hat{\setf{x}}_{i~,}^{k} \hat{\setf{x}}_j^{k}), 
					 \hat{f}^{k\minus1}(\setf{\hat{x}}_{i ~~,}^{k\minus1} \setf{\hat{x}}_j^{k\minus1}) \big\} 
	+ \hspace{-12pt}
	\sum_{f \in \Theta^{k} \setminus \Theta^{k\minus1}} \hspace{-12pt} \hat{f}^{k}(\setf{\hat{x}}_i^k,\setf{\hat{x}}_j^k) \notag \\
	\hspace{12pt} &\geq 
	\sum_{f \in \tilde{E}^k} \hat{f}^{k}(\setf{\hat{x}}_i^k,\setf{\hat{x}}_j^k) \\
	\hspace{12pt} &\geq 
	\sum_{f \in \tilde{E}^k} f(\setf{x}_i^*,\setf{x}_j^*). \tag{by Lemma~\ref{pr:ub}}
\end{align*}
Thus, combining the above it follows:
\begin{align*}
	& \hspace{-12pt}
	\sum_{f \in \Theta^{k\minus1} \setminus \tilde{E}^{k}} \hspace{-12pt} \hat{f}^{k\minus1}(\setf{\hat{x}}_i^{k\minus1},\setf{\hat{x}}_j^{k\minus1}) 
	+ \hspace{-12pt} 	\sum_{f \in \Theta^{k} \setminus \Theta^{k\minus1}} \hspace{-12pt} \hat{f}^{k}(\setf{\hat{x}}_i^k,\setf{\hat{x}}_j^k) \\	
	&\hspace{12pt} + \hspace{-12pt} 
	\sum_{f \in \Theta^{k\minus1} \cap \tilde{E}^k} \hspace{-12pt} 
		\max \big\{ \hat{f}^k(\hat{\setf{x}}_{i~,}^{k} \hat{\setf{x}}_j^{k}), 
					 \hat{f}^{k\minus1}(\setf{\hat{x}}_{i ~~,}^{k\minus1} \setf{\hat{x}}_j^{k\minus1}) \big\}  \notag \\
	\hspace{12pt} &\geq 
	\sum_{f \in \Theta^{k\minus1} \setminus \tilde{E}^{k}} \hspace{-12pt} 
	f(\setf{x}_{i}^{*}, \setf{x}_j^{*}) 
	+ \hspace{-12pt} 	
	\sum_{f \in \Theta^{k} \setminus \Theta^{k\minus1}} \hspace{-12pt} 
	f(\setf{x}_{i}^{*}, \setf{x}_j^{*}) \\
	&\hspace{12pt} + \hspace{-12pt} 
	\sum_{f \in \Theta^{k\minus1} \cap \tilde{E}^k} \hspace{-12pt} 
		\max \big\{ f(\setf{x}_{i}^{*}, \setf{x}_j^{*}), 
				     f(\setf{x}_{i}^{*}, \setf{x}_j^{*})) \big\}  \notag \\
&\geq \sum_{f \in \Theta^{k}} \hspace{-4pt} f(\setf{x}_i^*, \setf{x}_j^*).
\end{align*}
Which concludes the proof.
\end{sproof}.

Lemma 3 ensures that the utility associated to the functions optimized in the relaxed problems $\hat{P}$, up to iteration $k$, is an upper bound for the evaluation of the same set of functions, evaluated under the optimal solution for $P$. 
The above proof relies on the observation that the functions in $\Theta^k$ include exclusively those ones associated with the optimization of problems $\hat{P}^{\ell}$, with $\ell \leq k$, and that the functions over which the optimization process operates multiple times (in $\Theta^{k\minus1} \!\cap \tilde{E}^k$), are evaluated with their maximal value observed so far.

\begin{theorem}
	\label{th:ub}
	For each $LN^k$,
	$
		\hat{F}^k(\setf{\hat{x}}_k) \geq \mathcal{F}(\setf{x}^*).
	$
\end{theorem}
\begin{sproof} By definition of $\hat{F}^{k}(\setf{x})$, it follows that,
	\begin{align*}
		\hat{F}^{k}(\setf{x})
			&=  
			\sum_{f \in \mathcal{F}} \hat{f}^{k}(\setf{\hat{x}}_i^k,\setf{\hat{x}}_j^k)  \\
			&= \sum_{f \in \Theta^{k}} \hspace{-4pt} \hat{f}^{k}(\setf{\hat{x}}_i^k,\setf{\hat{x}}_j^k)  + 
				  \hspace{-4pt}
				 	\sum_{f \not\in \Theta^{k}} \hspace{-4pt} \hat{f}^{k}(\setf{\hat{x}}_i^k,\setf{\hat{x}}_j^k)  \\
			&=	\sum_{f \in \Theta^{k}} \hspace{-4pt} \hat{f}^{k}(\setf{\hat{x}}_i^k,\setf{\hat{x}}_j^k)  + 
				  \hspace{-4pt}
				 	\sum_{f \not\in \Theta^{k}} \hspace{-4pt} \max_{d_i, d_j} f(d_i, d_j) 
					\tag{by def.~of $\hat{f}^k$} \\
			&\geq
				\sum_{f \in \Theta^{k}} \hspace{-4pt} f(x_i^*, x_j^*) + 
			  \hspace{-4pt}
			 	\sum_{f \not\in \Theta^{k} } \hspace{-4pt} f(x_i^*, x_j^*)
				\tag{by Lemma~\ref{lm:ub}} \\
			&= \mathcal{F}(\setf{x}^*)
	\end{align*}
which concludes the proof.
\end{sproof}

\begin{figure*}[!t]
	\hspace{-2pt}
	\begin{minipage}[h]{0.99\linewidth}
	\center
	\includegraphics[width=0.95\linewidth]{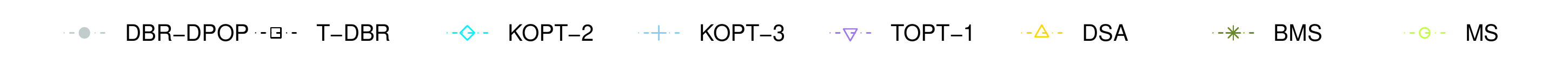}\\
 \includegraphics[width=0.31\linewidth,height=4cm]{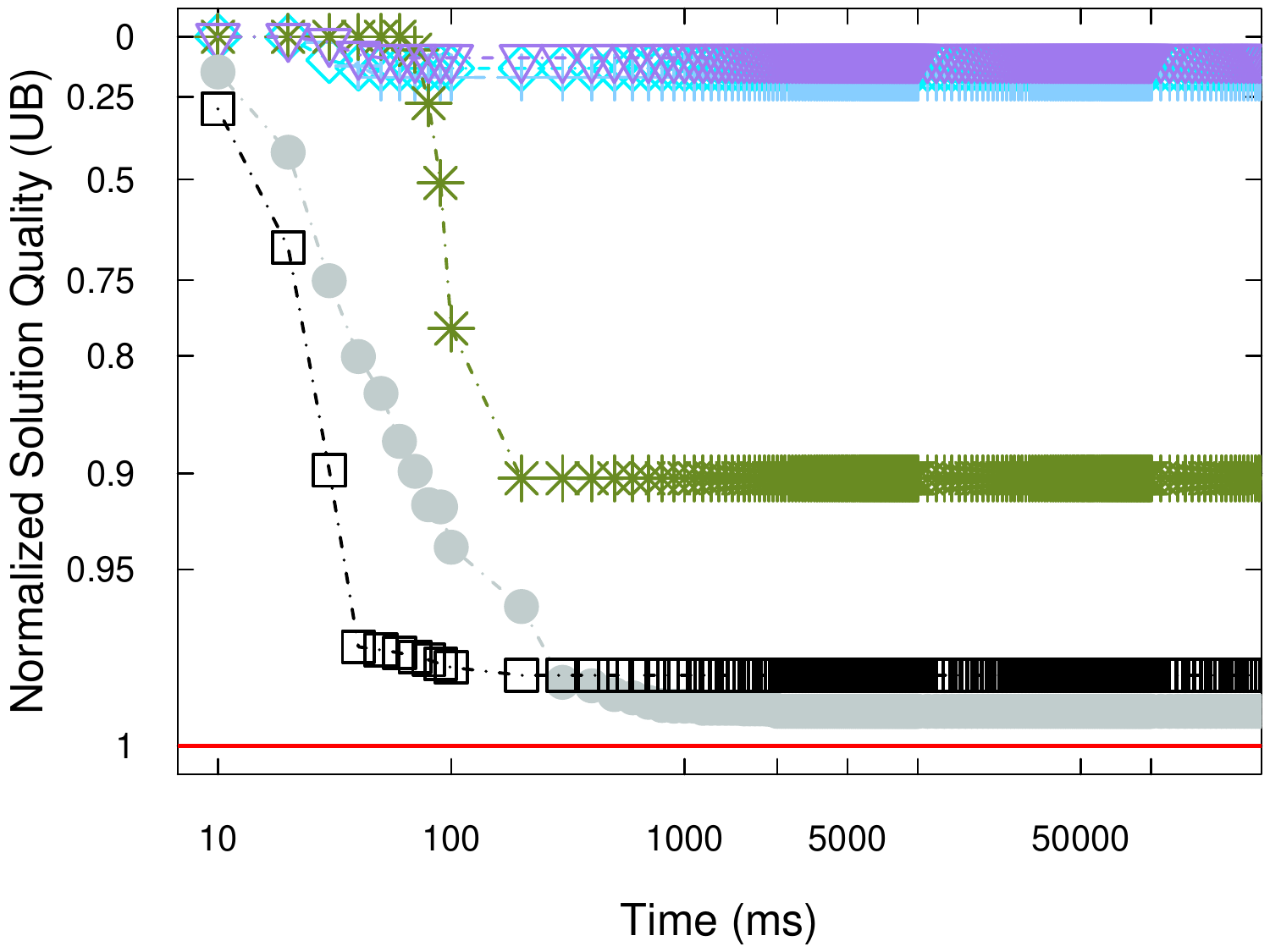}
  \hspace{5pt}
 \includegraphics[width=0.31\linewidth,height=4cm]{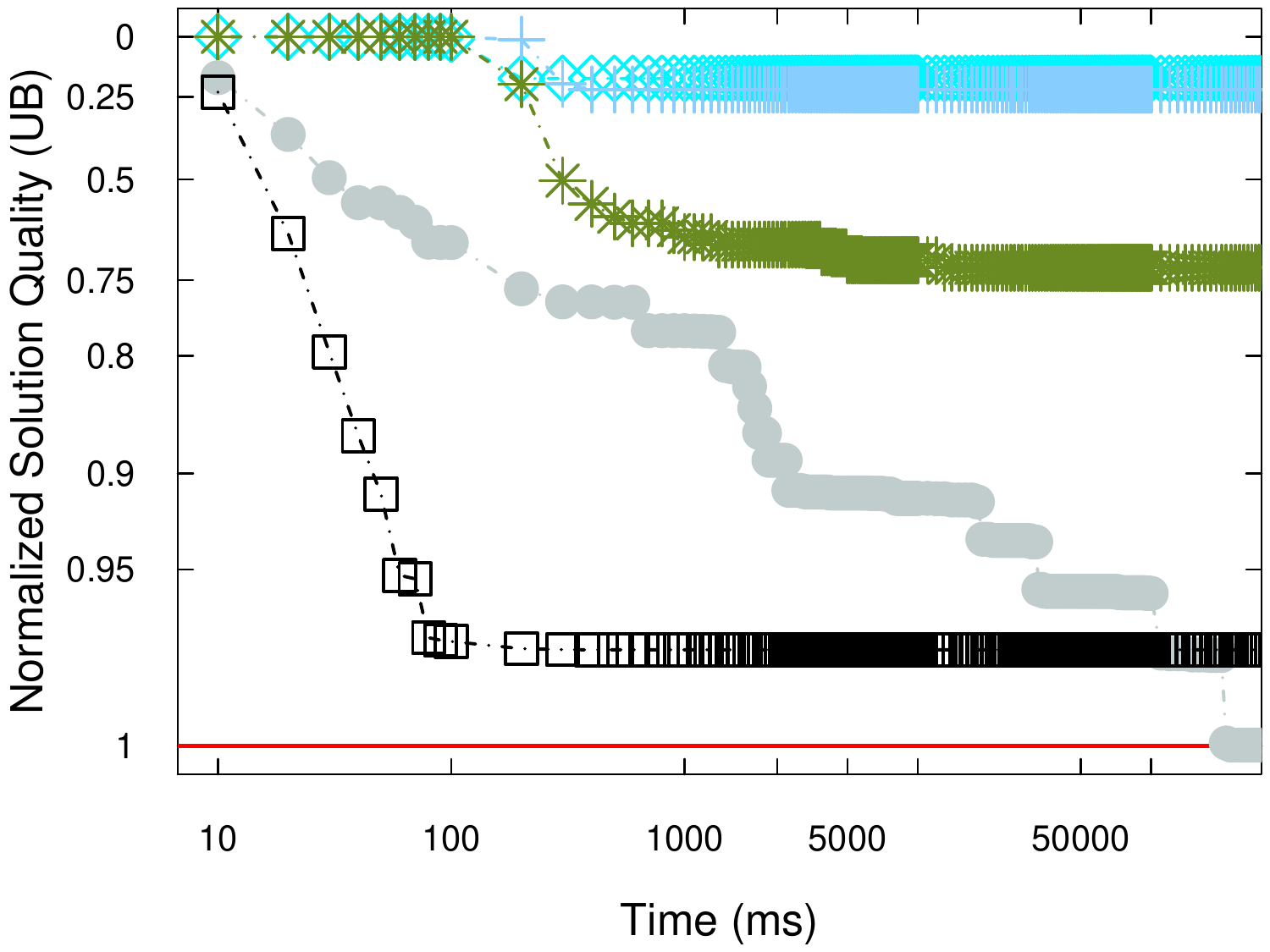}
  \hspace{5pt}
 \includegraphics[width=0.31\linewidth,height=4cm]{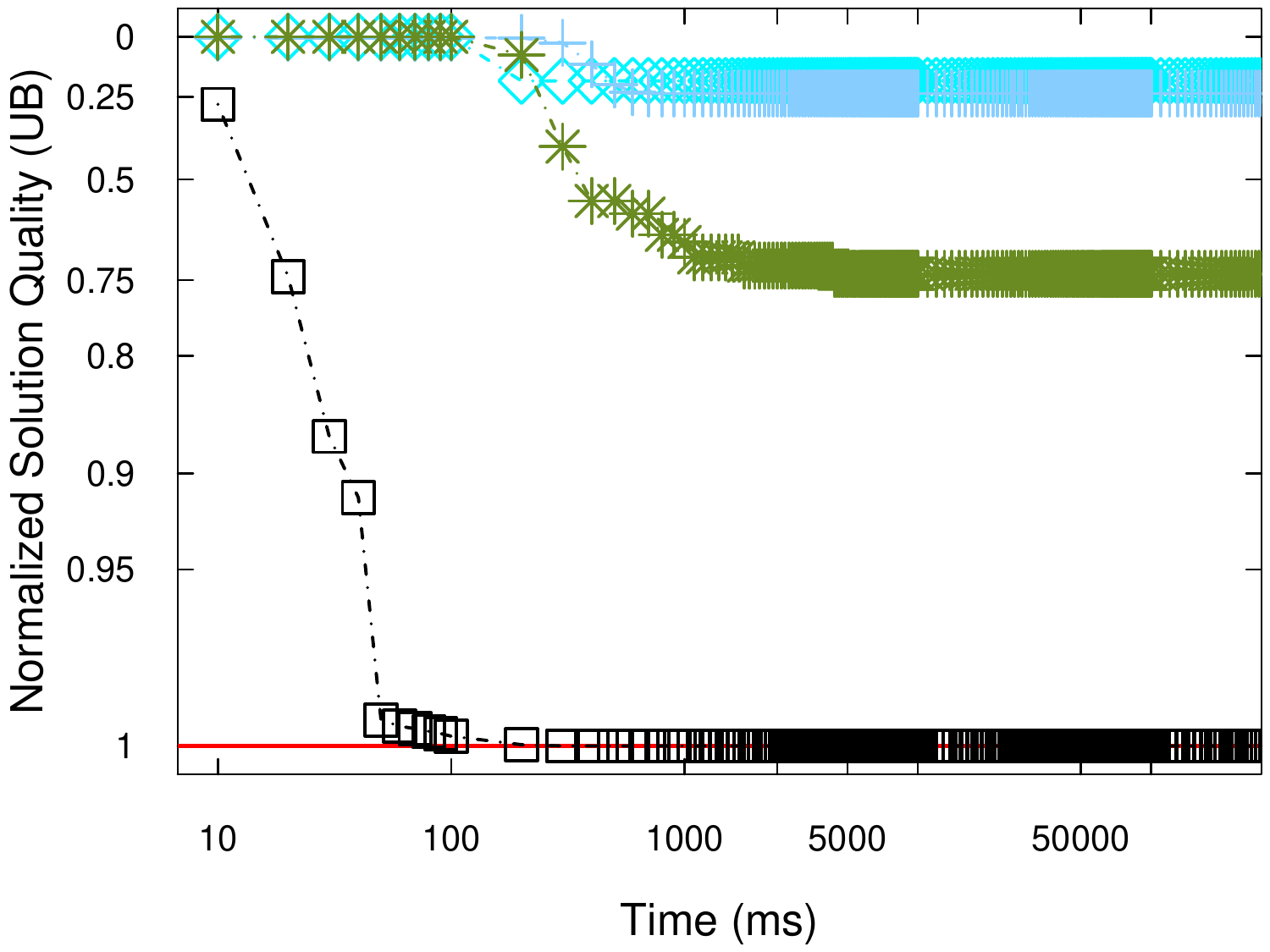}

  \includegraphics[width=0.31\linewidth,height=4cm]{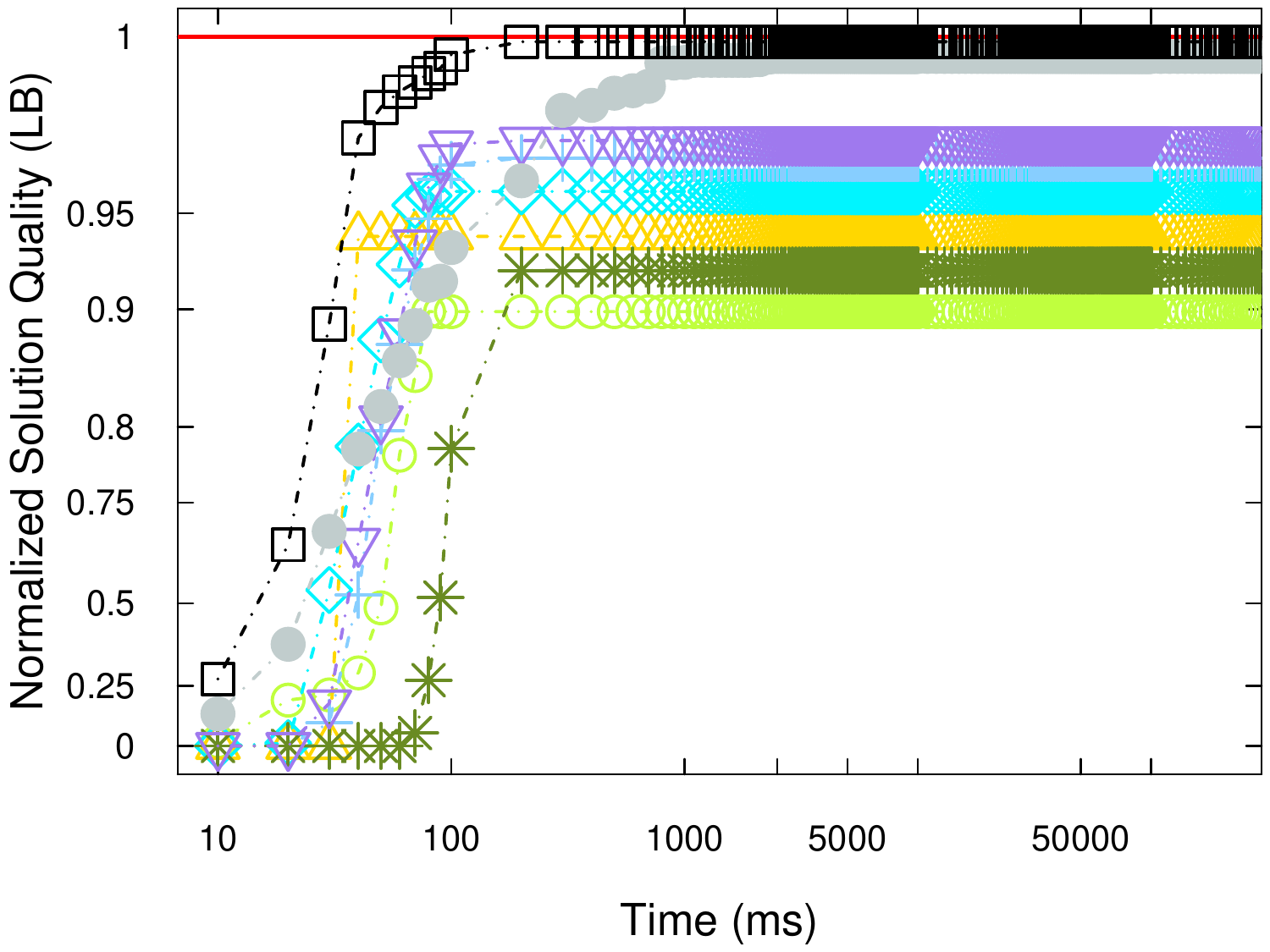}
  \hspace{5pt}
  \includegraphics[width=0.31\linewidth,height=4cm]{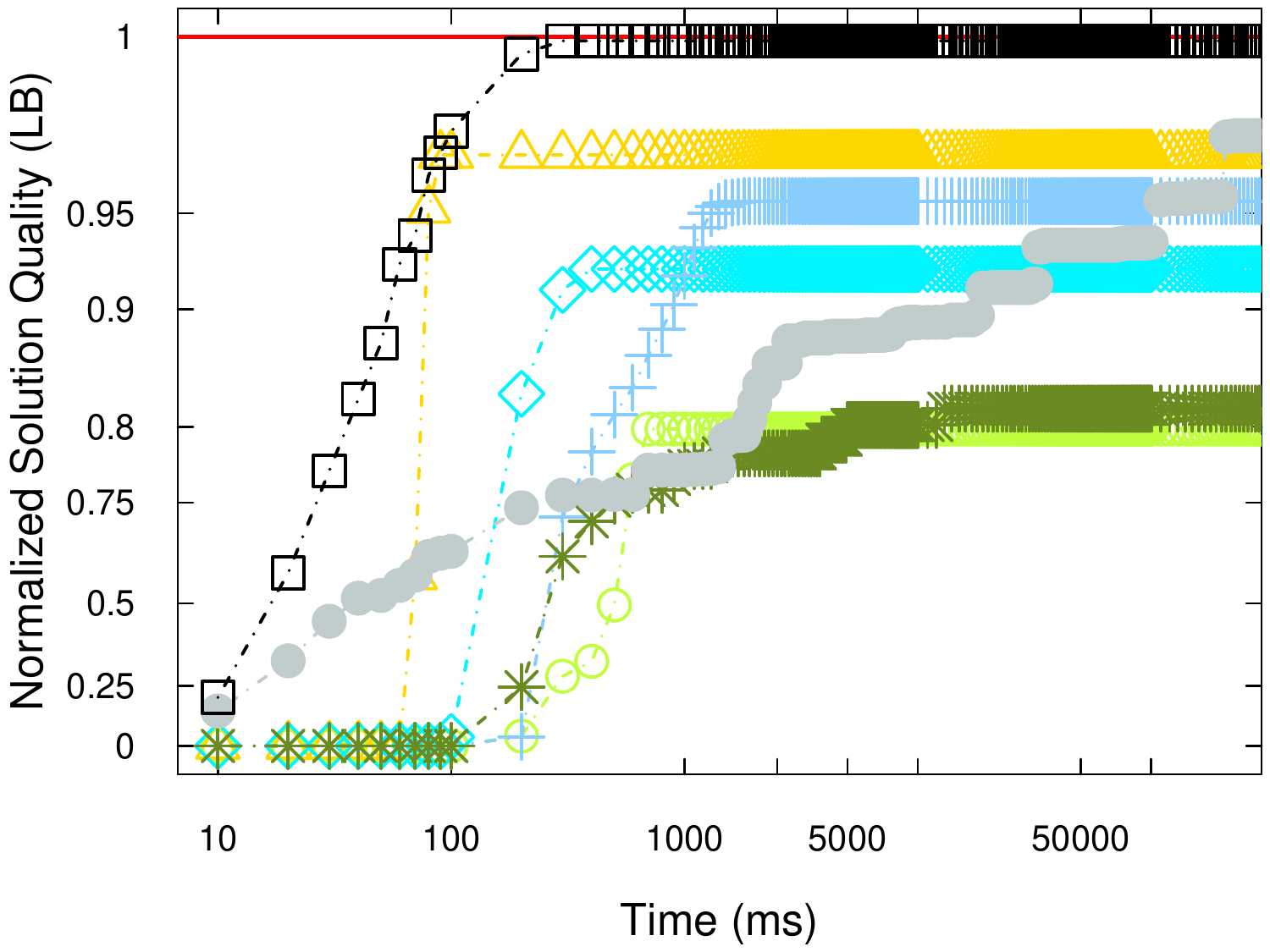}
  \hspace{5pt}
  \includegraphics[width=0.31\linewidth,height=4cm]{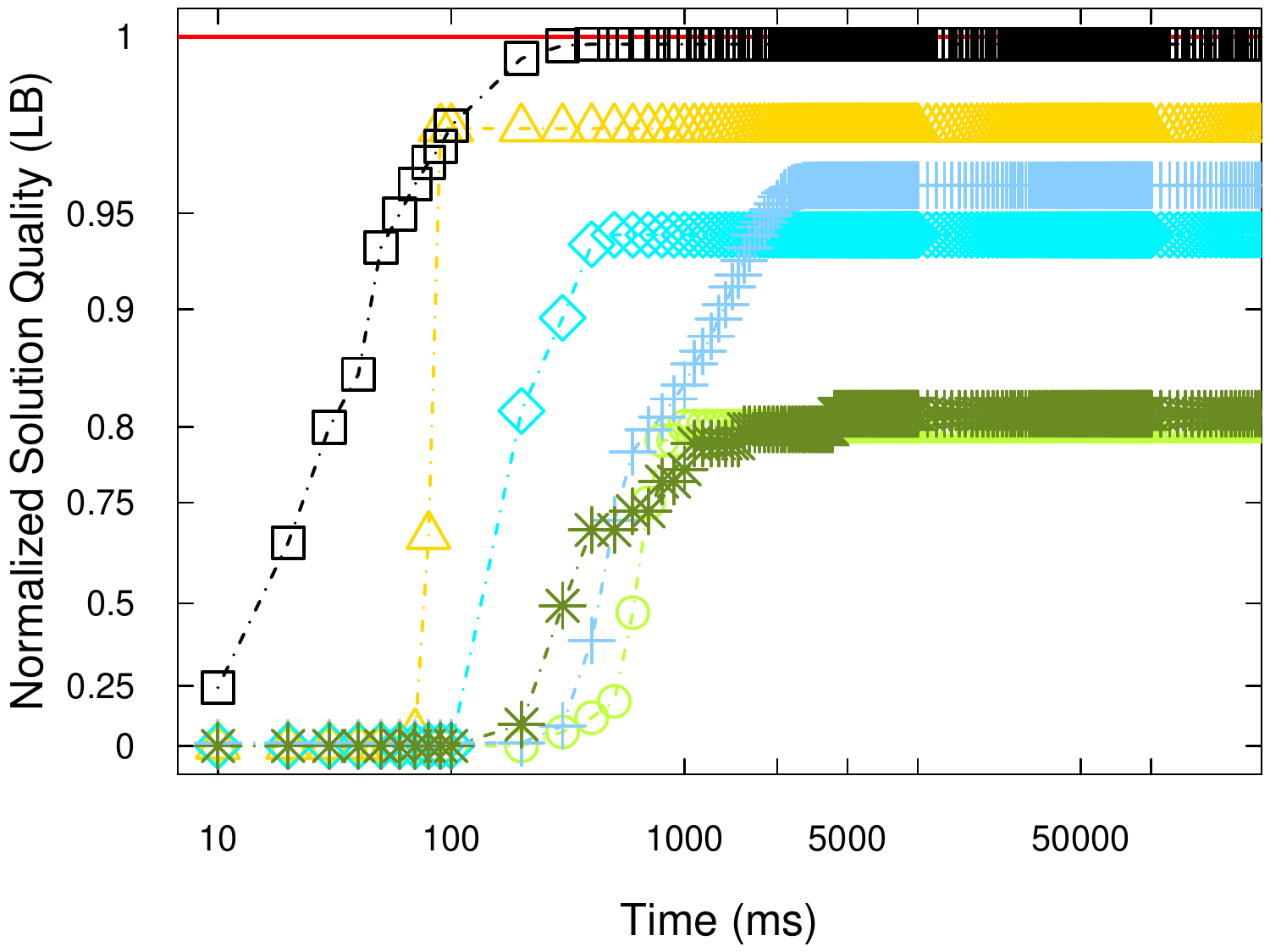}
  \end{minipage}
  \caption{Normalized solution quality for the upper bounds and lower bounds, on regular grids (left), random graphs (center), and scale-free (right) networks, at varying of the maximum time (top rows) and network load (bottom rows) allotted to the algorithms. 
  \label{fig:exp0}}
\end{figure*}

\begin{corollary}
An approximation ratio for the problem is
$$\rho = \frac{ \min_{k} \hat{F}^k(\hat{\setf{x}}^k) } { \max_k \mathcal{F}(\check{\setf{x}}^k) } 
\geq \frac{\mathcal{F}(\setf{x}^*)}{ \max_k \mathcal{F}(\check{\setf{x}}^k) }
$$ 
\end{corollary}
\begin{sproof}
This result follows from $\max_k \mathcal{F}(\check{\setf{x}}^k) \leq \mathcal{F}(\setf{x}^*)$ (Theorem~\ref{th:lb}) and $\min_k \hat{F}^k(\hat{\setf{x}}^k) \geq \mathcal{F}(\setf{x}^*)$ (Theorem~\ref{th:ub}). 
\end{sproof}

\smallskip

\begin{theorem}
In each iteration, T-DBR requires $O(\size{\mathcal{F}})$ number of messages of size $O(d)$, where $d = \displaystyle \max_{a_i \in \mathcal{A}} \size{D_i}$.
\end{theorem}
\begin{sproof}
	The number of messages required at each iteration is bounded by the \emph{Value Propagation Phase} of Algorithm~2, where each agent sends a message to each of its neighbors (lines 23 and 33). In contrast all other phases use up to $|\mathcal{A}|$ messages (which are reticulated from the leaves to the root of the pseudo-tree and vice-versa). The size of the messages is bounded by the \emph{Utility Propagation Phase}, where each agent (excluding the root agent) sends a message containing a value for each element of its domain (line 20).  All other messages exchanged contain two values (lines 23, 33, and 38). Thus the maximum size of the messages exchanged at each iteration is at most $d$.
\end{sproof}

\begin{theorem}
In each iteration, the number of constraint checks of each T-DBR agent is $O(d^2)$, where $d \is \displaystyle \max_{a_i \in \mathcal{A}} \size{D_i}$.
\end{theorem}
\begin{sproof}
	The number of constraint checks, performed by each agent in each iteration, is bounded by the operations performed during the Util-Propagation Phase.~In this phase, each agent (except the root agent) computes the lower and upper bound utilities  for each values of its variable $\setf{x}_i$ and its parent's variable $\setf{x}_{P_i^k}$ (lines~16--17).~
\end{sproof}

\section{Related Work}
\label{sec:related}

Aside from the incomplete algorithms described in the introduction, researchers have also developed extensions to complete algorithms that trade  solution quality for  faster runtime. For example, complete search algorithms have mechanisms that allow users to specify absolute or relative error bounds~\cite{modi:05,yeoh:09b}. 
Researchers have also worked on non-iterative versions of inference-based incomplete DCOP algorithms, with and without quality guarantees \cite{rogers:11,okimoto:11,petcu:05c}. Such methods are, however, unable to refine the initial solution returned.
Finally, the algorithm that is the most similar to ours is LS-DPOP~\cite{petcu:07c}, which operates on a pseudo-tree performing a local search. However, unlike D-LNS, LS-DPOP operates only in a single iteration, does not change its neighborhood, and does not provide quality guarantees. 

\section{Experimental Results}
\label{sec:results}

We evaluate the D-LNS framework against state-of-the-art incomplete DCOP algorithms, with and without quality guarantees, where we choose representative \emph{search}-, \emph{inference}-, and \emph{region optimal}-based solution approaches. 
We select Distributed Stochastic Algorithm (DSA) as a representative of an incomplete search-based DCOP algorithm; Max-Sum (MS), and Bounded Max-Sum (BMS), as representative of inference-based DCOP algorithms, and $k$- and $t$-optimal algorithms (KOPT, and TOPT), as representative of region optimal-based DCOP methods. All algorithms are selected based on their performance and popularity. 
We run the algorithms using the following implementations: 
We use the FRODO framework \cite{leaute:09} to run MS, and DSA,\footnote{We modified DSA-C in FRODO to DSA-B and set $p=0.6$.} 
the authors' code of BMS \cite{rogers:11}, and the DALO framework \cite{kiekintveld:10} for KOPT and TOPT. 
We systematically evaluate the runtime, solution quality and network load of the algorithms on binary constraint networks with \emph{random}, \emph{scale-free}, and \emph{grid} topologies, and we evaluate the ability of D-LNS to exploit domain knowledge over \emph{distributed meeting scheduling problems}.

\begin{table*}[!ht]
	\begin{center}
	\resizebox{\linewidth}{!}
	  {
	   \small \centering
	   \begin{tabular}{r|rrr|rrr|rrr|rrr|rrr|rrr|rr|rr|
	   }
		\cline{2-23}
		\multicolumn{1}{c|}{$|\mathcal A|$} 
	      & \multicolumn{3}{c}{DPOP-DBR}  
	      & \multicolumn{3}{c}{T-DBR} 
	      & \multicolumn{3}{c}{BMS} 
	      & \multicolumn{3}{c}{KOPT2} 
	      & \multicolumn{3}{c}{KOPT3} 
 	      & \multicolumn{3}{c}{TOPT1} 
	      & \multicolumn{2}{c}{MaxSum}  
	      & \multicolumn{2}{c|}{$\,\,$ DSA $\,\,$} \\
				& \multicolumn{1}{c}{$\rho$} & \multicolumn{1}{c}{$\epsilon$}  & \multicolumn{1}{c|}{$t$ (ms)} 
				& \multicolumn{1}{c}{$\rho$} & \multicolumn{1}{c}{$\epsilon$}  & \multicolumn{1}{c|}{$t$ (ms)} 
				& \multicolumn{1}{c}{$\rho$} & \multicolumn{1}{c}{$\epsilon$} & \multicolumn{1}{c|}{$t$ (ms)} 
				& \multicolumn{1}{c}{$\rho$} & \multicolumn{1}{c}{$\epsilon$} & \multicolumn{1}{c|}{$t$ (ms)} 
				& \multicolumn{1}{c}{$\rho$} & \multicolumn{1}{c}{$\epsilon$} & \multicolumn{1}{c|}{$t$ (ms)} 
				& \multicolumn{1}{c}{$\rho$} & \multicolumn{1}{c}{$\epsilon$} & \multicolumn{1}{c|}{$t$ (ms)} 
				& \multicolumn{1}{c}{$\epsilon$} & \multicolumn{1}{c|}{$t$ (ms)} 
				& \multicolumn{1}{c}{$\epsilon$} & \multicolumn{1}{c|}{$t$ (ms)}  \\
				\cline{2-23}	
	       10  & {\bf 1.06}	& {\bf 1.00}		& 2058 
	       	     & 1.15   & {\bf 1.00}  & 103
		     	     & 1.87  & 0.82		  & 211
			     & 4.33  & 0.94	  	& 63 
			     & 3.50  & 0.97	  	& 137 
			     & 6.00  & 0.99	  	& 998  	
			     & 0.78  				& 126 
			     & 0.94  				& {\bf 51}  \\
	       20  & {\bf 1.28} & 0.98 	& 58811 
	       		& 1.31       & {\bf 1.00} & 190
			& 2.30 & 0.82 			& 698
			& 7.67 & 0.92 			& 313 
			& 6.00 & 0.95 			& 1206
			& \oot{3}       		
			& 0.80    			& 441
			& 0.97 				& {\bf 69} \\
	       50  & \oot{3}		    		
	         	& {\bf 1.54} & {\bf 1.00} & 554 
			& 3.00 & 0.85   	& 2639
			& 17.66 & 0.90 		& 1961
			& 13.50 & 0.90 		& 8744
			& \oot{3}       	 	
			& 0.83  	  		& 2290
			& 0.99 				& {\bf 162} \\ 
	      100  & \oot{3}      		
	      		& {\bf 1.67} & {\bf 1.00} & 2101
			& 2.80 & 0.87 		& 27193
			& 34.33 & 0.90 		& 14884
			& 26.00 & 0.90 		& 86125
			& \oot{3}        		
			& 0.87  	  	& 5279
			& 0.98 			& {\bf 342} \\
	      200  & \oot{3}     	 		
	      		& {\bf 1.76} & {\bf 1.00} & 8990
			& 2.88 & 0.90 			& 37954 
			& 67.66 & 0.90 			& 217958
			& \oot{3}       		
			& \oot{3}        		
			&\oot{2} 
			& 0.97 				& {\bf 925} \\
	     \cline{2-23}
	  \end{tabular}
	  }
	\caption{Experimental results on \emph{random} networks \label{tab:exp1}}
\end{center}
\end{table*}

\noindent
The instances for each topology are generated as follows:

\bemph{Random:} We create an $n$-node network, whose density $p_1$ produces  $\lfloor n\,(n-1)\,p_1 \rfloor$ edges in total. We do not bound the tree-width, which is  based on the underlying graph.

\bemph{Scale-free:} We create an $n$-node network based on the Barabasi-Albert model~\cite{barabasi:99}. Starting from a connected $2$-node network, we repeatedly add a new node,  randomly connecting it to two existing nodes. In turn, these two nodes are selected with probabilities that are proportional to the numbers of their connected edges. The total number of edges is $2\,(n-2)+1$.

\bemph{Grid:} We create an $n$-node network arranged in a rectangular grid, where internal nodes are connected to four neighboring nodes and nodes on the edges (resp.~corners) are connected to two (resp.~three) neighbors.

We generate $50$ instances for each topology, ensuring that the underlying graph is connected. The utility functions are generated using random integer costs in $[0, 100]$. 
We set as default parameters, $\size{{\cal A}} \!=\! 20$, $|D_i| \!=\! 10$ for all variables, and $p_1 \!=\! 0.5$ for random networks.
We use a random destroy strategy for the D-LNS algorithms.
Algorithms' runtimes are measured using the \emph{simulated runtime} metric~\cite{sultanik:07}, and we impose a timeout of 300s. 
Results are averaged over all instances and are statistically significant\footnote{$t$-test performed with null hypothesis: DLNS-based algorithms find solution with better bounds than non-DLNS based ones.} with p-values $<0.0001$.
The experiment are performed  on an Intel i7 Quadcore 3.3GHz machine with 4GB of RAM. 

Figure~\ref{fig:exp0} illustrates the convergence results (normalized upper and lower bounds) for grids (left), random (center), and scale-free (right) networks in increasing amounts of maximum time allowed to the algorithms to complete. A value of 0 (1), means  worst (best) lower or upper bound w.r.t.~the lower or upper bound reported within the pool of algorithms examined.
All plots are in log-scale. 
These results show that the D-LNS-based algorithms converge to better solutions. In addition, they provide tighter upper bounds, and thus find better approximation ratios compared to the other algorithms.
The figures reporting the upper bounds do not illustrate MS and DSA, as they do not provide bounded solutions. 
TOPT-1 timed-out for all instances on random and scale-free networks. 
D-LNS with DPOP-DBR is slower than D-LNS with T-DBR, and it reaches a timeout for the scale-free networks. 
This is due to the fact that the complexity of the former repair phase is exponential in the induced width of the relaxed constraint graph, and  scale-free exhibit higher induced widths than grids and random network instances. In contrast, D-LNS with T-DBR does not encounter such limitations.
The main reason behind fast convergence to good solutions of the D-LNS algorithms is that, on average, about half of the agents are destroyed at each iteration, thus reducing the search space significantly. 
Additionally, the destroy phase of D-LNS is likely to create pseudo-forests, thus agents operating in different pseudo-trees can perform their operations concurrently.

Next, we validate our results at the varying of the number of agents in the problem, on random networks.
Table~\ref{tab:exp1} reports the approximation ratio $\rho$ and the ratio $\epsilon$ of the best quality found by all algorithms versus its quality, as well as the runtime $t$. Best approximation ratios, quality ratios, and runtimes are shown in bold.
The results show that D-LNS with DPOP-DBR finds better approximation ratios $\rho$ than those of the competing algorithms. However, it fails to solve problems bigger than $20$ agents. 
In contrast, D-LNS with T-DBR can scale to large problems better than other algorithms. 
Similarly to the trends observed in the previous experiment, DSA converges fastest to its solution for all problem sizes, however,
D-LNS with T-DBR finds better solutions w.r.t. all the other algorithms (i.e., better quality ratios $\epsilon$ and better approximation ratios $\rho$ for $|\mathcal{A}| > 20$).

\bigskip
\noindent{\textbf{Distributed Meeting Scheduling}}.
Many real-world problems model require the use of hard constraints,  to avoid considering infeasible solutions (see, e.g.,  {\small \url{http://www.csplib.org}}). 
We also evaluate the ability of our D-LNS framework to exploit such structure, exhibited in presence of domain-dependent knowledge and hard constraints, and test its behavior on \emph{distributed meeting scheduling problems}. In such problems, one wishes to schedule a set of events within a time range. We use the \emph{event as variable} formulation \cite{maheswaran:04b}, where events are modeled as decision variables. 
\begin{table}
	\resizebox{\linewidth}{!} {
	   \begin{tabular}{r |cc | cc | cc | }
		\multicolumn{1}{c}{Meetings:} 
	      & \multicolumn{2}{c}{20} & \multicolumn{2}{c}{50} & \multicolumn{2}{c}{100} \\
     		\cline{1-7}	
	      & \% SAT & TF (ms) & \% SAT & TF (ms)  & \% SAT & TF(ms)  \\
		\cline{2-7}	
	      DK  destroy
	           & {\bf 80.05}& {\bf 78}
	           & {\bf 54.11} &  {\bf 342}
	           & {\bf 31.20} & {\bf 718} \\
	      RN  destroy
	      	    & 12.45 & 648
	           & 1.00 & 52207
	           & 0.00 & -- \\
	      KOPT3
	      	    & 4.30 & 110367
	           & 0.00 & --
	           & -- & -- \\
	     \cline{2-7}
	  \end{tabular}
	  }
	\caption{Experimental results on \emph{meeting scheduling}. \label{tab:meet}}
\end{table}
Meeting participants can attend different meetings, and have time preferences that are taken into account in the problem formulation. Each variable can take on a value from the time slot range in $[0, 100]$, that is sufficiently early to schedule the required participant for the required amount of time. The problem requires that no meetings sharing some participants overlap.
 We generate the underlying constraint network using the random network model described earlier. The resulting meetings to schedule have 95, 613, and 2475 participants, in average respectively for the 20, 50, and 100 meetings experiments. We compare the repair phase T-DBR with both \emph{random} (RN) destroy and \emph{domain-specific knowledge} (DK) destroy methods. The latter destroys the set of variables in overlapping meetings. 
 Table \ref{tab:meet} reports the percentage of satisfied instances reported (\% SAT) and the time needed to find the first satisfiable solution (TF), averaged over $50$ runs.

The domain-specific destroy method has a clear advantage over the random one, being able to effectively exploit domain knowledge in presence of the hard constraints.
All other local search algorithm failed to report satisfiable solutions for any of the problems---only KOPT3 was able to find some satisfiable solutions for 20 meetings.

\section{Conclusions}
\label{sec:conclusions}

In this paper, we proposed a \emph{Distributed Large Neighborhood Search} (D-LNS) framework that can be used to find quality-bounded solutions in DCOPs. D-LNS is composed of a destroy phase, which selects a large neighborhood to search, and a repair phase, which performs the search over the selected neighborhood. 
We introduce two novel distributed repair phases, DPOP-DBR and T-DBR, built within the D-LNS framework, and characterized by low network usage; additionally, T-DBR provides a low computational complexity per agent.
Experimental results show that the D-LNS algorithms quickly converge to better solutions compared to incomplete DCOP algorithms that are representative of search-, inference-, and region-optimal-based approaches. 
The proposed results are significant---the ability of refining online quality guarantees, its quick convergence to good solutions, and the ability to exploit domain-dependent structure, makes D-LNS-based algorithms good candidates to solve a wide class of DCOP problems.
Additionally D-LNS can be extended to benefit of an anytime property, by using an anytime framework like that proposed in \cite{zivan:08}.
In the near future, we plan to investigate other schemes to incorporate into the repair phase of D-LNS, including constraints propagation techniques~\cite{bessiereGM12,fioretto:14b,gutierrezLLMM13} to better prune the search space, and techniques that actively exploit the bounds reported during the iterative procedure. 
We strongly believe that this framework has the potential to solve large distributed constraint optimization problems, with thousands of agents, variables, and constraints, and we plan 
to apply D-LNS based algorithms in the context of large distributed resource allocation problems in the near future.
 
\newpage
\bibliography{lns-dcop}


\end{document}